\documentclass[10pt,twocolumn,letterpaper]{article}
\usepackage{authblk}
\usepackage[pagenumbers]{cvpr} 

\usepackage{graphicx}
\usepackage{amsmath}
\usepackage{amssymb}
\usepackage{booktabs}
\usepackage{xcolor}
\usepackage{multirow}
\usepackage[normalem]{ulem}
\usepackage{subcaption}
\usepackage[pagebackref,breaklinks,colorlinks]{hyperref}

\usepackage[capitalize]{cleveref}
\crefname{section}{Sec.}{Secs.}
\Crefname{section}{Section}{Sections}
\Crefname{table}{Table}{Tables}
\crefname{table}{Tab.}{Tabs.}

\def\cvprPaperID{6025} 
\def\confName{CVPR}
\def\confYear{2022}

\makeatletter \renewcommand\AB@affilsepx{\hfill \protect\Affilfont} \makeatother
\begin{document}
\title{Long-term Visual Map Sparsification with Heterogeneous GNN}
\author[1]{Ming-Fang Chang}
\author[2]{Yipu Zhao}
\author[2]{Rajvi Shah}
\author[2]{Jakob J. Engel}
\author[1]{Michael Kaess}
\author[3]{Simon Lucey}
\affil[1]{Carnegie Mellon University}
\affil[2]{Meta Reality Labs Research}
\affil[3]{The University of Adelaide}

\twocolumn[{%
\maketitle
\renewcommand\twocolumn[1][]{#1}%
\begin{center}
	\centering
	\vspace{-7mm}
	\includegraphics[height=0.25\textwidth]{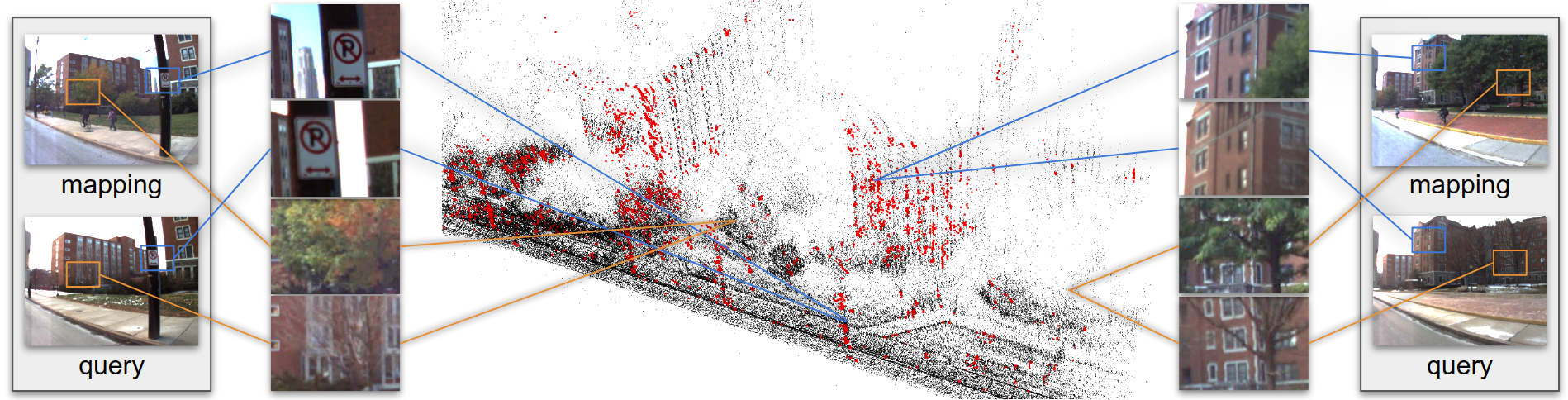}
	\captionof{figure}{
	Given a map built from SfM
, our proposed approach leverages GNNs and is able to identifies map points on stable structures (red points and blue squares), while discarding points that are prone to seasonal change, such as tree foliage (black points and orange squares).} 
	\label{fig:front}
\end{center}}
]

\begin{abstract}
\vspace{-3mm}

We address the problem of map sparsification for long-term visual localization. For map sparsification, a commonly employed assumption is that the pre-build map and the later captured localization query are consistent. 
However, this assumption can be easily violated in the dynamic world. Additionally, the map size grows as new data accumulate through time, causing large data overhead in the long term. 
In this paper, we aim to overcome the environmental changes and reduce the map size at the same time by selecting points that are valuable to future localization.
Inspired by the recent progress in Graph Neural Network (GNN), we propose the first work that models SfM maps as heterogeneous graphs and predicts 3D point importance scores with a GNN, which enables us to directly exploit the rich information in the SfM map graph. Two novel supervisions are proposed: 
1) a data-fitting term for selecting valuable points to future localization based on training queries; 
2) a K-Cover term for selecting sparse points with full-map coverage. The experiments show that our method selected map points on stable and widely visible structures and outperformed baselines in localization performance.

\vspace{-5mm}


\end{abstract}

\section{Introduction}
\label{sec:intro}



In long-term visual localization, a common strategy is to build and accumulate maps from the captured image streams, and then localize new incoming queries by matching against the accumulated map. 
In the presence of environmental changes, the accumulated map contains an increasing number of points and many of which are outdated. 
This will affect both the computational cost and the performance of localization in the long run. 
Therefore, the ability to identify and remove these invalid points is important for many applications that target dynamic environments, such as autonomous driving, field robotics, and Augmented Reality. 
Additionally, for devices with limited on-board memory, it enables keeping a compact map that only contains the most valuable information for future localization queries.

Existing works on map sparsification mostly fall into the category of subset selection, i.e., treating the 3D map as an over-sampled representation of a static world and aiming to select the most valuable point subset from them. The selection of point subset is typically formulated as a K-Cover problem. 
Assuming the map keyframes cover all the possible camera positions, the K-Cover algorithm encourages each keyframe in the map to observe K points under a total point number constraint~\cite{lynen2019d, dymczyk2015, park2013,mera-trujillo2020a}. 
These methods are purely based on the historical data stored in the map, therefore lacking the ability to identify points invalidated due to environmental changes. 
When the environment changes, the map can only be updated by collecting new query data over the whole mapped area and solve the K-Cover problem again with the new query data, which is inefficient and expensive.
Apart from sparsifying a 3D map, there are some works on selecting 2D key points, e.g., by predicting the persistency \cite{dymczyk2016will} or the repeatability \cite{doan2021learning} of visual features.
However, the predictors proposed only take instantaneous measurements (such as local image patches) and not exploit the full context stored in the accumulated map. 


Recently, Graph Neural Networks (GNN) have shown promising results with data with different structures, such as citation graphs~\cite{velickovic2018}, local feature matching
~\cite{sarlin2020} and visibility graphs~\cite{shen2021}. In this work, we exploit this flexibility of GNNs to formulate map sparsification as a learning problem and overcome the limitations of previous methods. First, by modeling the SfM map as a graph, we can directly employ the context-rich SfM map as the GNN input instead of instantaneous measurements. Second, in contrast to the K-Cover based methods that requires full-extent new queries to update the map, we are able to train a GNN with only partial queries and use it to sparsify the whole map. A main improvement from previous methods is the ability to incorporate the partial new data and select important points from the whole map according to the partial new data, as there is no trivial way for the baseline methods to do this without collecting new data that covers the whole mapped area.

To this end, we propose the first work that extracts features from SfM maps with a heterogeneous GNN. 
We first represent the SfM map with a heterogeneous graph, where 3D points, 2D key points and images are modeled as graph nodes, and the context such as the visibility between 2D and 3D points are modeled as graph edges. Afterwards, we use a heterogeneous GNN to predict map point importance scores based on the local appearance and the spatial context in the map graph. 
In addition, we propose two novel losses to guide the training: 
1) a data-fitting term that selects points based on the appearance  and the spatial distribution of the training query data, and 2) a K-Cover loss term that drives to sparse point selection with full-map coverage. 
When evaluated on an outdoor long-term dataset with significant environmental changes (Extended CMU Seasons \cite{sattler2018a}), our approach can select map points on stable and widely-visible structures (e.g., buildings/utility poles), while discarding points on changing object (e.g., foliage) or with highly repetitive texture (e.g., pavement). 
Compared with the K-Cover baseline \cite{lynen2019d}, our approach outperforms in visual localization performance with the same map size.

\vspace{-2mm}
\section{Related Works}

In this section, we first briefly describe the literature of robust feature learning, then review the existing map sparsification works, and finally cover relevant studies on GNNs that inspired our work.

\subsection{Robust Feature Learning}

Many previous works have attempted to solve the long-term visual localization problem by finding robust feature descriptors against environmental changes~\cite{toft2020long} 
(such as day-night, lighting conditions, and seasonal changes). 
Concrete examples include R2D2~\cite{revaud2019}, SOSNet~\cite{tian2019sosnet}, PixLoc~\cite{sarlin2021back} and \cite{iccv2021challenge}. 
Some methods look into the dynamics of visual features (and the corresponding physical environment) such as persistency~\cite{dymczyk2016will} and repeatability~\cite{doan2021learning}. Besides learning robust features, some works also attempt to overcome the environmental challenges by finding common information in 2D and 3D, such as semantic information~\cite{toft12017, toft12018 } and predicting depth from query images~\cite{piasco2019}.
In this work, instead of finding robust features, we focus on sparsifying the SfM map globally by taking the whole map graph structure into consideration. We use Kapture \cite{humenberger2020}, a modern mapping and localization library using R2D2, to generate data and evaluate the proposed method.



\subsection{Map Sparsification}
For a map that contains redundant information of a world, the goal of map sparsification is to select the most valuable subset. In previous works, it is common to assume that the map contains all the possible camera positions, and formulate the map compression as a K-Cover problem, which encourages each possible camera position (the key frame location in the map) to observe enough 3D points for performing robust PnP during localization under a total point number budget.
The K-Cover problem is then solved using various techniques: a probabilistic approach~\cite{cao2014}, Integer Linear Programming (ILP)~\cite{ lynen2019d, dymczyk2015} and Integer Quadratic Programming (IQP)~\cite{park2013,dymczyk2015,mera-trujillo2020a}. A hybrid map and hand-crafted heuristics were also used to determine the importance of map points~\cite{camposeco2019, Luthardt2018, muhlfellner2016}. These methods work well in a static world but suffer from performance degradation in vastly dynamic environments where many of the visibility edges in the map are outdated and invalidated.

\subsection{Graph Neural Networks}
Graph Neural Networks (GNNs) \cite{hamilton2018} have been applied to a variety of learning tasks with irregular data structures, such as citation graphs \cite{velickovic2018} and image visibility graphs \cite{shen2021}.
An important advantage of Graph Neural Networks is the ability to handle heterogeneous data \cite{wang2021a}. 
In this work, we represent the various information in SfM maps with heterogeneous graphs and extract features with a GNN.
Recently, attention-based networks have shown strong performance in feature extraction from not only sequential data~\cite{vaswani2017} but also graph structures such as 2D-3D matching \cite{sarlin2020}. Inspired by these works, we investigate the combination of heterogeneous GNN and attention, and demonstrated better final performance than the baselines.


\section{Approach}

\begin{figure}
    \centering
	\includegraphics[width=0.35\textwidth]{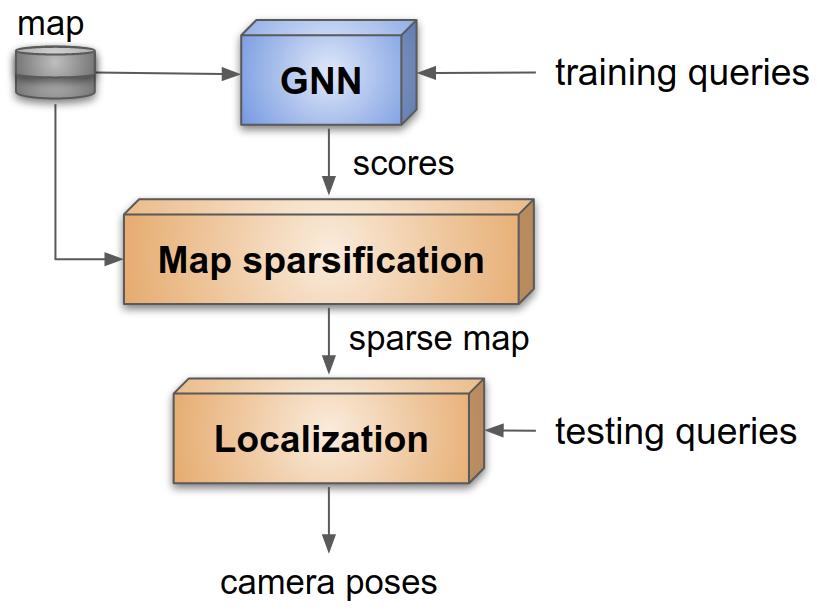} 
	\caption{Overall framework. The proposed GNN learns to predict a score for each 3D point in the map. The predicted scores are used to sparsify the map. 
 We report the performance of localizing a set of testing queries to the sparsified map. 
	}
	\label{fig:system_block}
	\vspace{-4mm}
\end{figure}

Given an SfM map and a set of localization queries recorded at different times in a large-scale dynamic environment, our goal is to select a subset of 3D map points that are most informative, i.e. result in high localization performance. 
To achieve this, we first turn the input SfM map into a heterogeneous graph (Sec.~\ref{sec:graph}) and train an attention-based GNN (Sec.~\ref{sec:gatconv}, \ref{sec:gnn}) to predict the importance scores for 3D map points, which are then used to sparsify the map.
Finally, we localize the testing query set against the sparsified map, and report the localization performance (Sec.~\ref{sec:eval}). An illustration of our overall system flow is shown in Fig.~\ref{fig:system_block}.


\begin{figure*}[h]
	\begin{tabular}{c c c c}
		\includegraphics[height=0.2\textwidth]{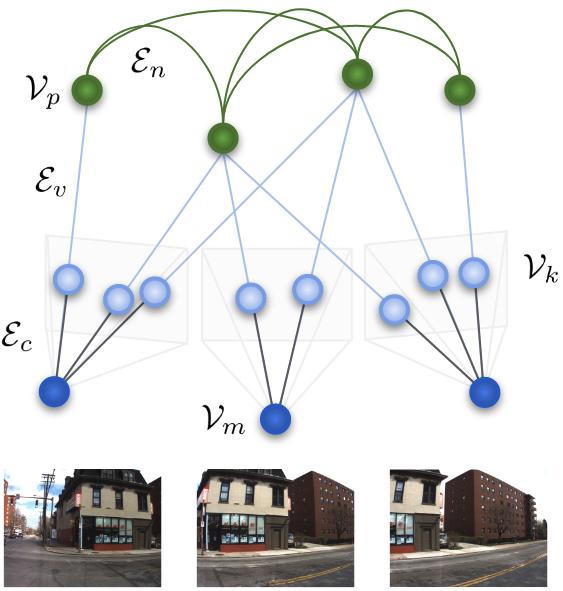}  & \includegraphics[height=0.2\textwidth]{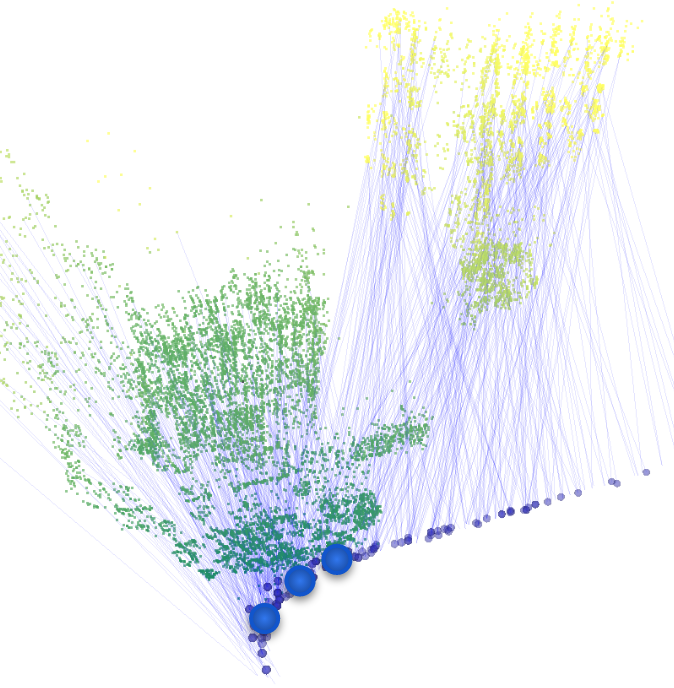}  &  
		\includegraphics[height=0.2\textwidth]{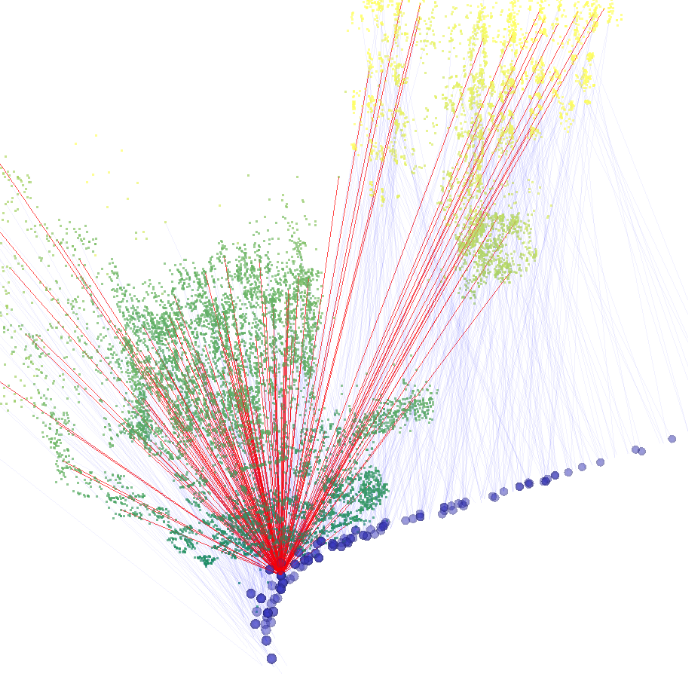} &
		\includegraphics[height=0.2\textwidth]{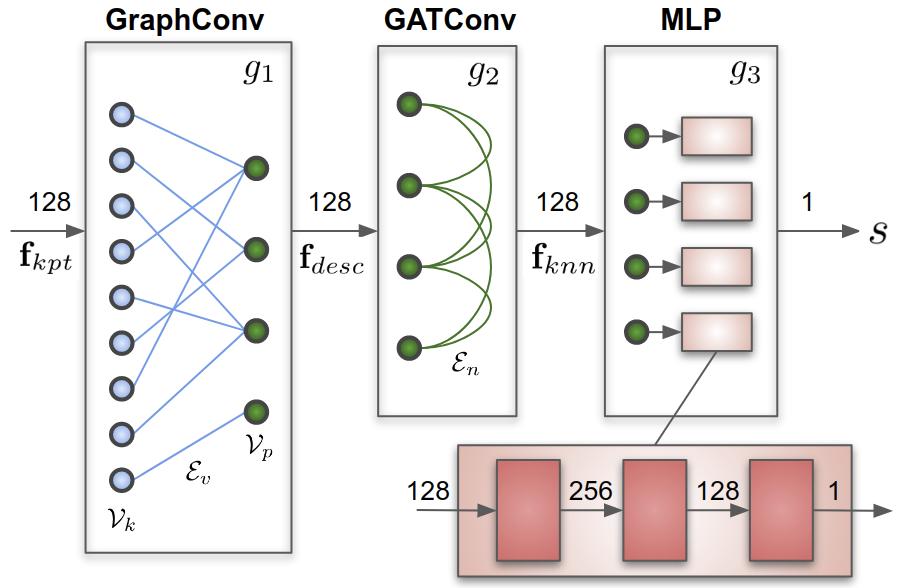} \\
		{\small(a)} & {\small(b)} & {\small(c)} & {\small(d)}
	\end{tabular}	
	\caption{An SfM map as a heterogeneous graph and the network structure. a) A simplified graph: dark blue circles are image nodes $\mathcal{V}_m$, light blue circles are key point nodes $\mathcal{V}_k$, and green circles are 3D point nodes $\mathcal{V}_p$. The edges $\mathcal{E}_c$, $\mathcal{E}_v$, and $\mathcal{E}_n$ are containing edges, visibility edges, and kNN edges, represented by black, light blue, and green colors. (b) A real snapshot of the Extended CMU Seasons dataset. Image nodes $\mathcal{V}_m$ and visibility edges $\mathcal{E}_v$ are as blue dots and lines. The key point nodes $\mathcal{V}_k$ are not shown. The color on the 3D points $\mathcal{V}_p$ encodes the distance to the current query image with green being low values and yellow high values. Three image node positions corresponding to the images in (a) are labeled with with dark blue circles. (c) In each training iteration, we sample an image node and trace the corresponding edges to extract a subgraph to run our GNN. The $\mathcal{E}_v$ used to extract this subgraph are shown as red lines. (d) Our network takes the key point descriptors $\mathbf{f}_{kpt}$ and predicts a score $s$ for each map point. We define three network layers: $g_1$ that aggregates descriptors to 3D points, $g_2$ that collects 3D local information, and $g_3$ as the final per-point MLP (pink blocks). A dark pink block is an MLP layer, which contains a linear layer and a LeakyReLU activation. The numbers above the arrows are feature dimensions.}
	\label{fig:graphs}

\end{figure*}

\subsection{SfM Map as Heterogeneous Graph}
\label{sec:graph}
A heterogeneous graph by definition is a graph structure that contains different types of nodes or edges. To represent an SfM map, three types of nodes are defined: 3D point nodes $\mathcal{V}_p$, 2D key point nodes $\mathcal{V}_k$, and image nodes $\mathcal{V}_m$. We also define three types of edges: visibility edges $\mathcal{E}_v$ connecting corresponding $\mathcal{V}_p$ and $\mathcal{V}_k$, kNN edges $\mathcal{E}_n$ connecting each $\mathcal{V}_p$ and its k nearest neighboring $\mathcal{V}_p$, and containing edges $\mathcal{E}_c$ connecting each $\mathcal{V}_k$ to the corresponding image $\mathcal{V}_m$.  Each $\mathcal{V}_p$ might be connected to multiple $\mathcal{E}_v$s and $\mathcal{V}_k$s because it is observed by multiple map images. The SfM map is then represented with a heterogeneous graph $\mathcal{G}=\{\mathcal{V}_p, \mathcal{V}_k, \mathcal{V}_m , \mathcal{E}_v, \mathcal{E}_n, \mathcal{E}_c\}$. An illustration of our map graph is shown in Fig.~\ref{fig:graphs}(a)(b).

The per-point importance score is predicted based on local appearance and spatial context. 
We design our map graph to provide the information: 
first, the local appearance data are stored in $\mathcal{V}_k$ by embedding the key point descriptors extracted at the map building stage. 
Second, the spatial context is captured in kNN edges $\mathcal{E}_n$, which are derived from the 3D point positions stored in $\mathcal{V}_p$. 
The image nodes $\mathcal{V}_m$ do not carry features, but are used to trace connected $\mathcal{V}_k$ and $\mathcal{V}_p$ for ensuring the GNN selects enough number of $\mathcal{V}_p$ in the field-of-view of each $\mathcal{V}_m$, as shown in Fig.~\ref{fig:graphs}(c).

In practice, 
we store two sets of $\mathcal{V}_k$, $\mathcal{V}_m$, $\mathcal{E}_v$, and $\mathcal{E}_c$ in the map graph: 
one set is from the map and the other set is from localizing the query set on the map before sparsification. 
The first set is fed to the proposed GNN to provide information for score prediction. The second set is only available in the training area, and was only used to generate the point selection labels $L_{gt}$ stored in $\mathcal{V}_p$ (Sec.~\ref{sec:losses}).

Note that all the graph edges described above are directional. To be specific, $\mathcal{E}^{ji}_n$ represents a kNN edge from a neighbor $\mathcal{V}_p^j$ to the $\mathcal{V}_p^i$, and $\mathcal{E}^{wi}_v$ shows a visibility edge from key point $\mathcal{V}_k^w$ to map point $\mathcal{V}_p^i$, where $i,j,w$ are node indices. The directionality of edges is useful in retrieving local subgraphs during network training (Sec.~\ref{sec:gnn}).

\subsection{Graph Attention Network}
\label{sec:gatconv}

To extract the spatial context from the map, we propose to aggregate the features from locally connected 3D point nodes with a Graph Attention Network (GATConv)~\cite{wang2020c, velickovic2018}. 
For a 3D point node $\mathcal{V}_p^i$, a GATConv layer is applied to fuse the input node features and predict an output node feature. 
Formally, the GATConv operation is:
\begin{equation}
	\begin{aligned}
	\alpha^h_{ij} &= softmax_j (a( \mathbf{W}^h\mathbf{h}_i , \mathbf{W}^h\mathbf{h}_j))\\
	\mathbf{h}_i^{+} &= \sum_{h=1}^H\sum_{j\in \{1,\dots, k+1\}} \alpha^h_{ij} \mathbf{W}^h\mathbf{h}_j ,
	\end{aligned}
	\label{eq:gatconv}
\end{equation}
where $\mathbf{h}_j \in \mathbb{R}^{F}$ is an input feature from $\mathcal{V}_p^j$ to node $\mathcal{V}_p^i$ with feature dimension $F$. The input features are from the $\mathcal{V}_p^i$ itself and the kNN nodes, where $j \in \{1,2,\dots,k, i\}$ and $k$ is the number of kNN nodes. 
The $\mathbf{W}^h\in \mathbb{R}^{F^+ \times F}$ is a shared weight matrix, $\alpha^h_{ij}$ is the normalized attention coefficient, $H$ is the number of attention heads, $a(.):  \mathbb{R}^{F^+} \times \mathbb{R}^{F^+} \rightarrow \mathbb{R}$ computes the attention coefficients. We aggregate the multi-head GATConv outputs by simple summation. 
The output $\mathbf{h}_i^{+} \in \mathbb{R}^{F^+}$ is the output feature with dimension $F^+$ stored on $\mathcal{V}_p^i$. 
Empirically, we found this GATConv outperformed GraphConv\cite{kipf2017} and SAGEConv~\cite{hamilton2018} for our application.

\subsection{Heterogeneous Graph Neural Network}
\label{sec:gnn}

We design a heterogeneous GNN to extract features and perform score prediction from the aforementioned map graph. The motivation is that the key point descriptors, although not raw pixel values, still contain valuable appearance information, enabling us to infer the 3D point scores from the connected 2D key point descriptors. The heterogeneity here enables us to define different operations according to the node and edge types. 

Our GNN comprises three stages: 1) a descriptor gathering layer $g_1$, 2) a local feature extraction layer $g_2$, and 3) a final Multilayer Perceptron (MLP) layer $g_3$. In $g_1$, we trace the connected $\mathcal{E}_v$ for each $\mathcal{V}_p$ to collect the connected key point descriptors stored in $\mathcal{V}_k$. The collected descriptors are sent to a Graph Convolutional layer (GraphConv)~\cite{kipf2017} with LeakyReLU activation and summation aggregation functions. The output of $g_1$ is an aggregated point feature $\mathbf{f}_{desc}$ carrying the local appearance information. In $g_2$, we use the GATConv layer (Sec.~\ref{sec:gatconv}) to gather the nearby point features from the kNN $\mathcal{V}_p$, generating a local feature $\mathbf{f}_{knn}$ that captures spatial context. Finally, a 3-layer MLP $g_3$ is used to convert the point feature dimension to 1 and a sigmoid layer is used to constrain the predicted score value $s$ to $[0,1]$. The network structure is shown in Fig.~\ref{fig:graphs}(d). 

Let $i,j \in \{1,2, 
\dots, N_p\}$ denote the map point indices and $w \in \{1,2, \dots, N_k\}$ be a key point index, where $N_p$ and $N_k$ are the total number of map points and key points. Let $\mathcal{G}$ denote the map graph, the score prediction steps are:
\begin{equation}
	\begin{aligned}
	\mathbf{f}^i_{desc} &= g_1 ( \{ \mathcal{V}^w_k | \mathcal{E}^{wi}_v \in \mathcal{G} \})\\
	\mathbf{f}^i_{knn} &= g_2 ( \{ \mathbf{f}^j_{desc} |   \mathcal{E}^{ji}_n \in \mathcal{G} \} )\\
	s^i &= Sigmoid (g_3 (\mathbf{f}^i_{knn} )),\\
	\end{aligned}
\end{equation}
where $\mathbf{h}_i =\mathbf{f}_{desc}^i$ and $\mathbf{h}_i^+ =\mathbf{f}_{knn}^i$ in Eq.~\ref{eq:gatconv}.

To facilitate GNN training on large-scale graphs, we sample a $\mathcal{V}_m$ to extract a local subgraph for each training batch and only run our GNN on the local subgraph. 
Given a $\mathcal{V}_m$, we first extract the connected $\mathcal{V}_k$ by tracing $\mathcal{E}_c$. Afterwards we trace $\mathcal{E}_v$ and $\mathcal{E}_n$ to extract the corresponding $\mathcal{V}_p^i$ and its neighbors. Finally, we trace the  $\mathcal{E}_v$ connecting to the neighboring $\mathcal{V}_p^j$ for computing the neighboring $\mathbf{f}^j_{desc}$.

\subsection{Training Losses}
\label{sec:losses}

Our losses promote high scores on points with two properties: 
first, the descriptor distribution of the selected points should \emph{align with the descriptors that are useful for training query localization}.
Second, the selected points should \emph{cover all the possible viewing poses}, so that all the queries would observe a sufficient amount of points within the field-of-view. 
We propose a training loss with two terms: 

\noindent\textbf{Data Fitting Term.} Since the ILP baseline performs well in a static environment~\cite{lynen2019d}, we use it as an oracle to generate point selection labels. We first localize the training queries on the map, collecting the 2D-3D matches between the training queries and the map, and run the ILP baseline~\cite{lynen2019d} to obtain the point selection results, which is a binary vector $L_{gt}$. The ILP baseline in this setting, denoted as ILP (query), factors out the environmental changes and performs well (Fig.~\ref{fig:result_curves}(a)), but cannot be achieved in the real world unless the training queries cover the whole mapped area.
The data fitting term is then computed by comparing the predicted scores $\mathbf{S}$ and  $L_{gt}$ with a Binary Cross Entropy (BCE) loss $\mathcal{L}_{BCE}$:
\begin{equation}
	\begin{aligned}
		\mathcal{L}_{BCE} = BCE(L_{gt}, \mathbf{S}).
	\end{aligned}
\end{equation}


For the maps we evaluated with, we found the computation of ILP formulation is tractable to process the whole map. It is also possible to use IQP~\cite{park2013} for label generation, but in practice IQP is computationally intractable to run on large-scale maps without additional graph partition steps. The potential effect of graph partition on localization performance is beyond the focus of this paper.

\noindent\textbf{K-Cover Term}. Training the network with $\mathcal{L}_{BCE}$ alone would only encourage point selection that aligns with $L_{gt}$ in the training set, but it does not guarantee map point coverage across the whole map. 
To compensate this, we leverage transductive learning and additionally encourage the sum of the scores of all the $\mathcal{V}_p$ connected to each $\mathcal{V}_m$ to be close to a predefined positive integer K, which indicates the number of 3D points each image should observe to support robust localization. Empirically, we observed that this setup converges faster during training than the case not penalizing the samples larger than K. Upon satisfying the K-Cover constraint, we also encourage the score sparsity to select fewer points with an $L_1$ norm loss. Letting $l$ be the index of image node $\mathcal{V}_m$, we define $\phi_l$ as the set of map point indices that selects the set of $\mathcal{V}_p$ whose connected $\mathcal{V}_k$ is within $\mathcal{V}_m^l$ (as the red edges in Fig.~\ref{fig:graphs}(c)). The score prediction of $\mathcal{V}_p^i$ is denoted as $s_i$. The final K-Cover loss is:
\begin{equation}
	\begin{aligned}
		\phi_l &= {\{ i |   \mathcal{E}^{lw}_c \in \mathcal{G} \cap  \mathcal{E}^{wi}_v \in \mathcal{G} \}},\\
		\mathcal{L}_{KC} &= \sum_l|K - \sum_{i \in \phi_l} s_{i}| + \lambda||\mathbf{S}||_{1}.
	\end{aligned}
\end{equation}

By adding both terms, we propose the final loss as:
\begin{equation}
	\begin{aligned}
		\mathcal{L} = \mathcal{L}_{BCE} + \mathcal{L}_{KC}.
	\end{aligned}
\end{equation}

The data split and usage is summarized in Tab.~\ref{tab:split}. Note that the training and testing queries are spatially non-overlapping and the pre-built map covers both the train and test areas. The role of the training queries is to provide up-to-date appearance information that cannot be obtained from the outdated map data, as we focus on the temporal appearance difference. In this case, the training and testing data should not overlap spatially but can overlap temporally.

\section{Evaluation}
\label{sec:eval}
In this section, we describe the data preparation process, implementation details and experimental results.

\begin{figure*} [h]
	\centering
	\begin{tabular}{c c}
		\includegraphics[width=0.46\textwidth]{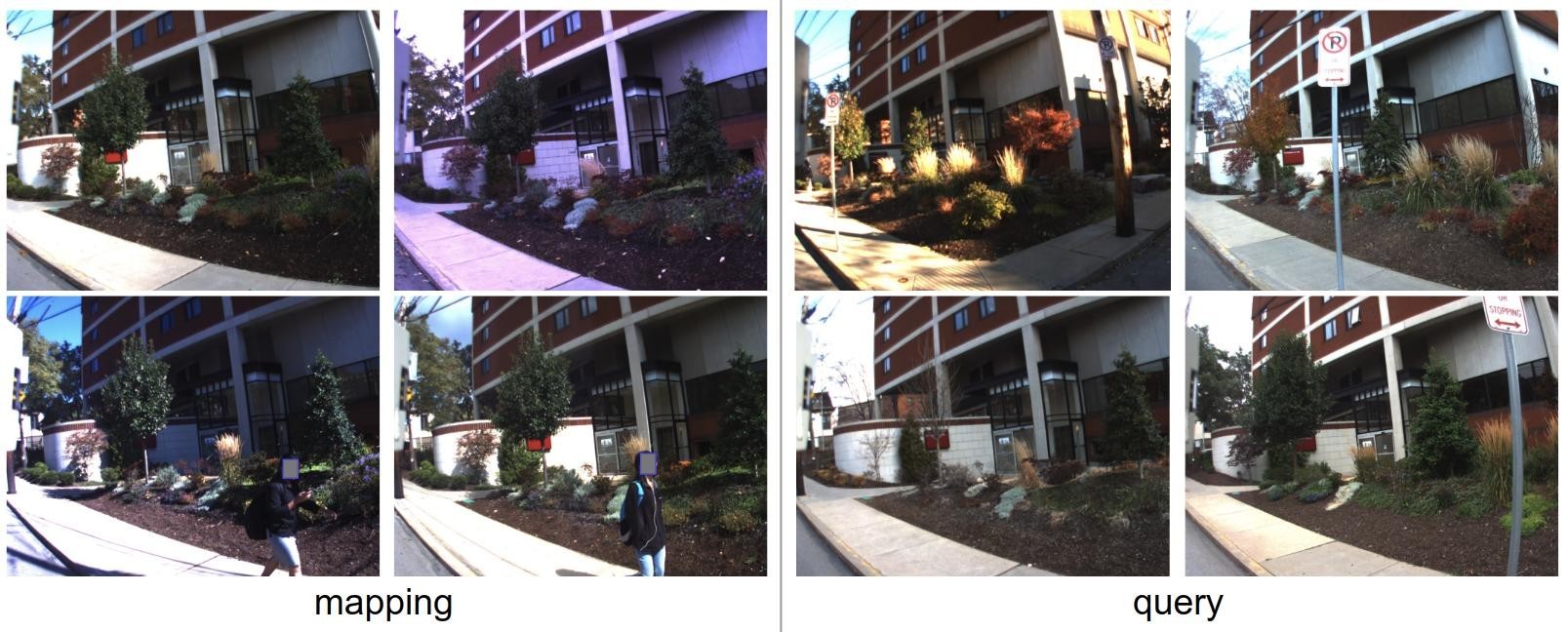}  & \includegraphics[width=0.46\textwidth]{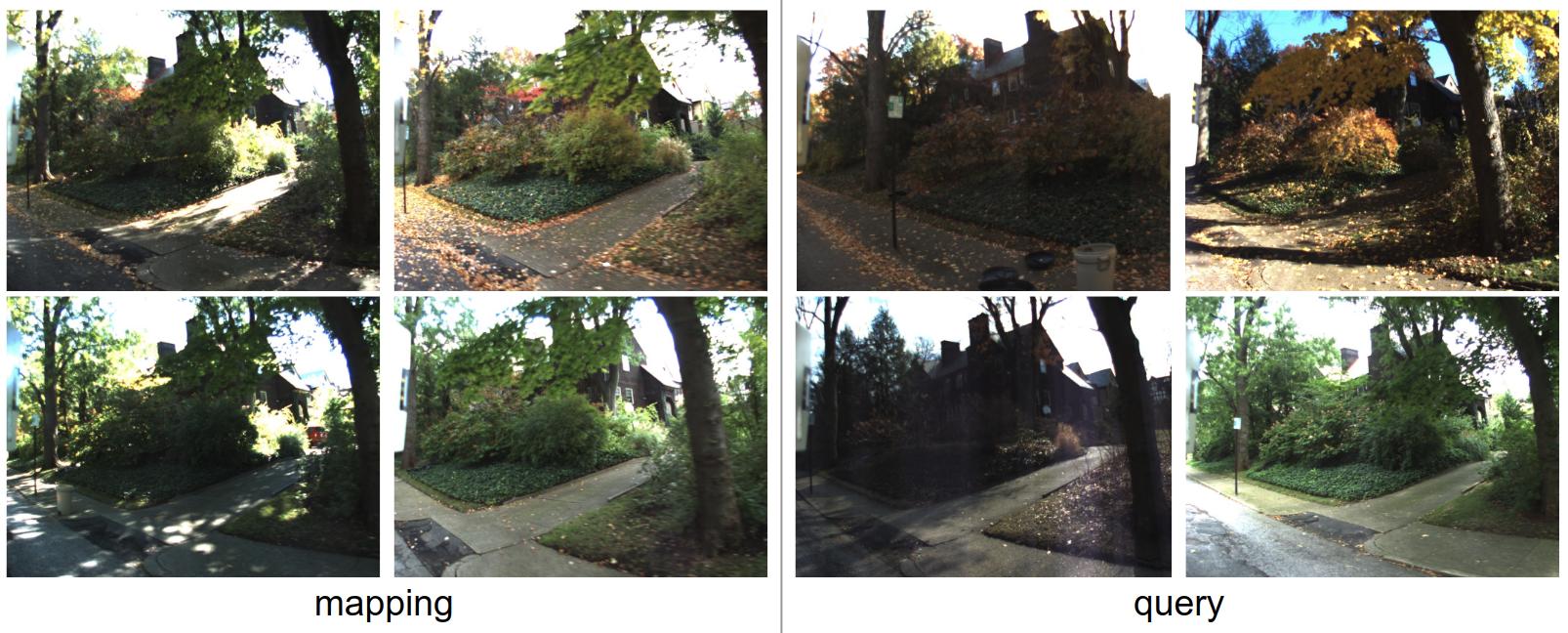}  \\
		{\small(a) Example images from slice 3} & {\small(b) Example images from  slice 11 }\\
		\includegraphics[width=0.46\textwidth]{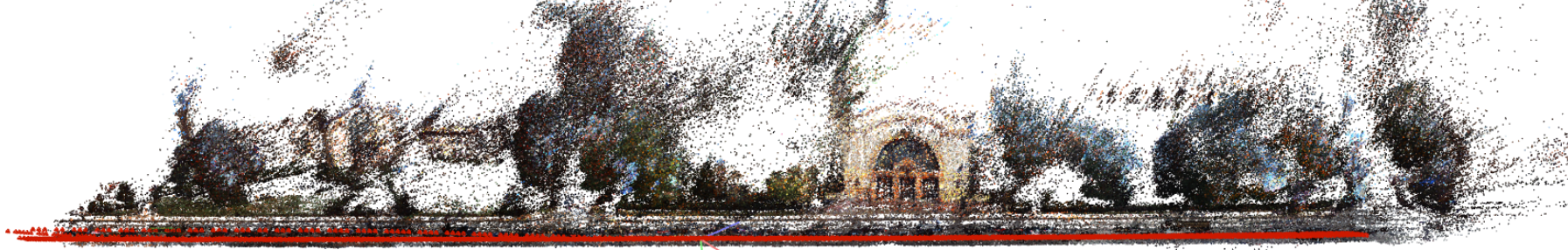}  & \includegraphics[width=0.46\textwidth]{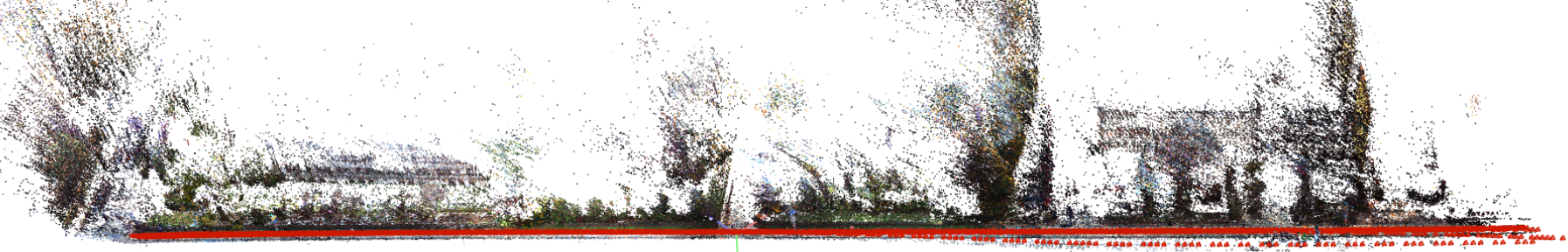} \\
		{\small(c) Training set map example (camera 0 of slice 4)} & {\small(d) Test set map example (camera 1 of slice 4)}

	\end{tabular}	
	\caption{Example images from Extended CMU Seasons dataset. We observed large seasonal changes across the whole dataset. In (a)(b), on the left are the map image examples and on the right are query image examples recorded at similar locations. In addition, the Extended CMU Seasons dataset was recorded by two cameras. We used camera 0 (c) for training and camera 1 (d) for validation/testing. The training and test sets capture two sides of the road with no spatial overlap. The red dots at the bottom are the map image locations. }
	\label{fig:datasets}
\end{figure*}
\noindent\textbf{Data Preparation}
We evaluated our approach on Extended CMU Seasons dataset~\cite{sattler2018a,Badino2011}, which consists of 12 sessions recorded by two cameras across months in multiple locations. To simulate the natural accumulation of map data, we used sessions 0-5 to build a multi-session map, and used sessions 6-11 as the query set to localize. The mapping and query sets have significantly different appearance. The map was built with Kapture~\cite{humenberger2020}. The localization performance was measured by registering the query sets on the multi-session maps built from session 0-5 . We used 13 slices (scenes) for evaluation, including the Urban and Suburban slices (3-4, 6-16), and discarded the Park slices and slice 2, 5 due to poor localization performance on the raw multi-session map before sparsification. The 13 slices for evaluation contained various objects such as vegetation, buildings, and moving objects, and multiple weathers like sunny, cloudy, and snowy. An example of seasonal appearance changes is shown in Fig.~\ref{fig:datasets}(a)(b).

We further split the query set by the two cameras (camera 0, camera 1), and used camera 0 of all the 13 slices for training, the camera 1 of slice 3 for validation, and the camera 1 of the other 12 slices for testing. The number of mapping/query images in each data set split are 17837/16077 for training, 1333/1428 for validation, and 16498/15627 for testing. Note that the camera 0 and camera 1 point towards two sides of the road and have no overlap as Fig.~\ref{fig:datasets}(c)(d). 

\begin{table}
    \caption{The data splits by type and usage. There are two cameras in the Extended CMU Seasons dataset, noted by c0 and c1. We separated the 12 sessions temporally and used the old sessions (0-5) for mapping, the new sessions as queries (6-11).}
    \centering
    \footnotesize
    \begin{tabular}{c|cc|cc|c}
    \toprule
     \multirow{2}{*}{\bf{Data Type}} & \multicolumn{2}{c|}{\bf{Spatial}} & \multicolumn{2}{c|}{\bf{Temporal}} &\multirow{2}{*}{\bf{Used for}}\\\cline{2-5}
     & train & test & old & new\\
    \midrule
    map ($\mathcal{G})$ & \checkmark (c0) & \checkmark (c1) & \checkmark & & $\mathcal{L}_{KC}$, $\mathcal{L}_{BCE}$\\
    query (train) & \checkmark (c0) & & & \checkmark &  $\mathcal{L}_{BCE}$\\
    query (test) &  & \checkmark (c1) & & \checkmark & not used \\
    \bottomrule
    \end{tabular}
    \label{tab:split}
\end{table}
\noindent\textbf{Implementation Details}
The proposed GNN is implemented with PyTorch and Deep Graph Library (DGL) \cite{wang2020c}. During the training process, we loop through the map image nodes $\mathcal{V}_m$ in the training set to extract subgraphs to run GNN on. A four-layer DGL node sampler ($\mathcal{V}_m \leftarrow \mathcal{V}_k \leftarrow \mathcal{V}_p \leftarrow~knn~\mathcal{V}_p \leftarrow \mathcal{V}_k~of~knn~\mathcal{V}_p$) was used to extract the subgraph in each training iteration to provide necessary information. It took about 3.97s to process a map graph (with average $4.12\times10^5$ map points) on an Nvidia Quadro RTX 3000 GPU and an i7-10850H CPU @ 2.70GHz. More graph statistics are in the appendix. 

As for parameters, we used $k=9$ to build kNN edges among 3D points, $K=30$ and $\lambda = 0.01$ in the K-Cover loss. 
The ILPs~\cite{lynen2019d} were implemented using Gurobi\cite{gurobioptimizationllc2021}, and is configured with $b = 30$. We used $n_{desired} = 500$~\cite{lynen2019d} to generate $L_{gt}$.
The network was trained with an AdamW optimizer with learning rate $0.001$ and $\beta$s $(0.9, 0.999)$ for 20 epochs. For each case, we selected the epoch with the best validation performance for testing.

Final evaluation is conducted with the Kapture localization pipeline~\cite{humenberger2020}. Given a query image, it first retrieves the map images with similar global features, and then performs 2D-2D key point descriptor matching between the query image and the retrieved map images. The 3D points corresponding to the matched map key points are used to perform PnP with the matched query key points.
The Kapture default R2D2~\cite{revaud2019} descriptor is used in map building, localization, and as our network input $\mathbf{f}_{kpt}$. 

\subsection{Localization Performance on Sparsified Maps}

For each map sparsification method, we first obtained its point selection result, and reconstructed the multi-session map in Kapture format with only the the key points and descriptors that correspond to the selected points. We used the number of point descriptors remaining in the map (\#kpts) as map size proxy, since these high-dimension descriptors (e.g., 128 for R2D2) occupied most of the map storage space. Three baselines were compared:

\begin{itemize}
    \item \textbf{Random} : randomly select a subset of map points up to the allowed budget.
    \item \textbf{ILP (map)}  : the conventional ILP~\cite{lynen2019d}, which assembles the K-Cover problem with 1) the visibility edges stored in the map, and 2) the per-point weight based on number of observations in the map.
    \item \textbf{ILP (query)} : the ideal ILP~\cite{lynen2019d} that has access to test queries. The K-Cover problem is constructed using visibility edges from localizing the test queries on the map before sparsification, and points are weighted according to the number of observations during the test query localization. This approach indicates the ideal performance of ILP approach without environmental changes and cannot be achieved in the real world.
\end{itemize}




We obtained data points by sweeping the desired total point number $n_{desired}$\cite{lynen2019d}. For our method, we randomly selected points with predicted scores larger than 0.1. If there were not enough points with scores larger than 0.1 to satisfy $n_{desired}$, we randomly selected from the rest of the points. We observed that predicted score distribution is close to binary (due to the $L_1$ norm sparsity loss) and the point selection result is not sensitive to the score threshold.

\begin{figure*}
	\centering
	\begin{tabular}{c c}
		 \includegraphics[height=0.29\textwidth]{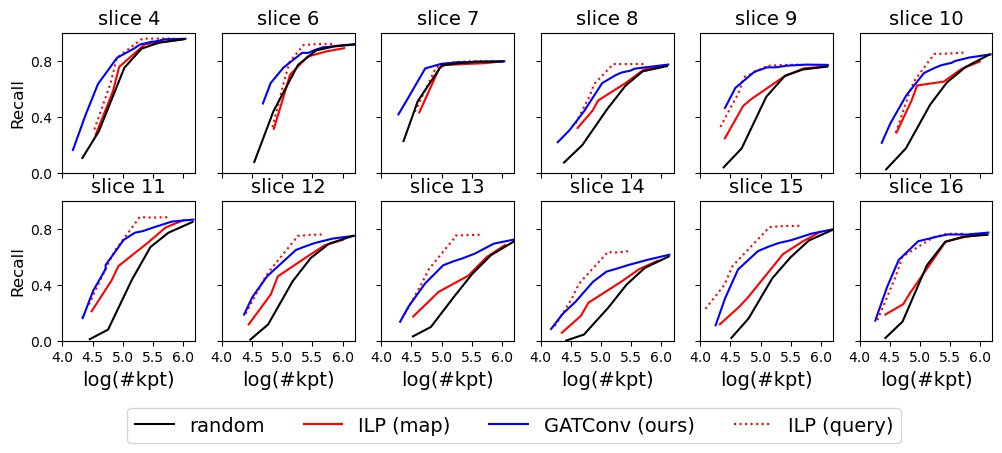}  & \includegraphics[height=0.29\textwidth]{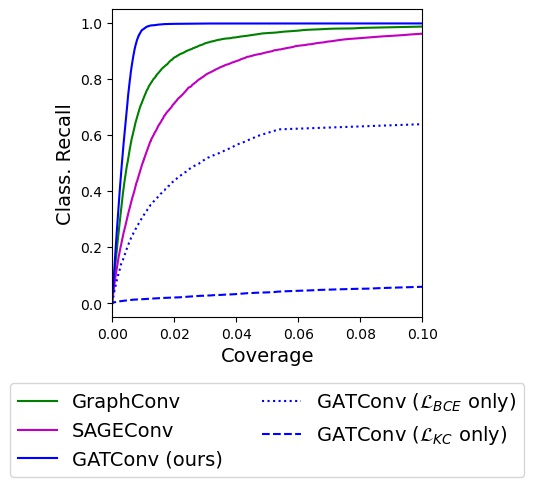}   \\
		{\small(a) The recall vs. map size curves for each slice in the test set} & {\small(b) Classification performance}
	\end{tabular}
	\vspace{-3mm}
	\caption{ Localization and classification recall comparisons. (a) Our approach outperformed the ILP (map) and the random baselines in all test slices, achieving higher recalls (success rate) under the same map size budgets. On the other hand, the ILP (query) also significantly outperformed ILP (map), showing the impact of environmental changes on baselines. The recall error thresholds here are 0.25m and 2.0$^\circ$. (b) Compared with ILP (query), the GATConv trained with the full proposed loss achieved the highest classification recall (ratio of selected positive labels) under the same coverage (ratio of the number of selected points against total number of points). }
	\label{fig:result_curves}
	\vspace{-4mm}
\end{figure*}

\begin{figure*} [h]
	\centering
	\begin{tabular}{c c}
		\includegraphics[width=0.44\textwidth]{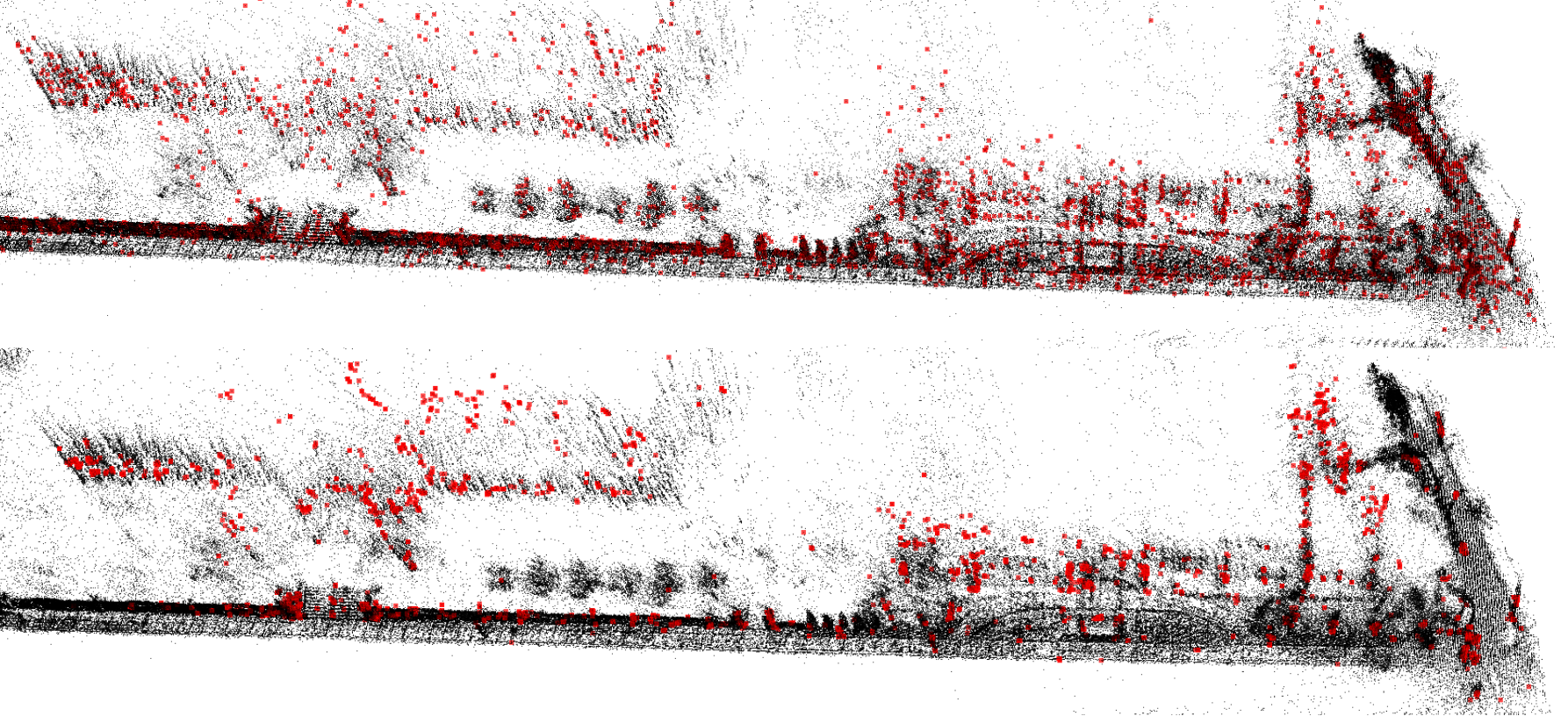}  & \includegraphics[width=0.43\textwidth]{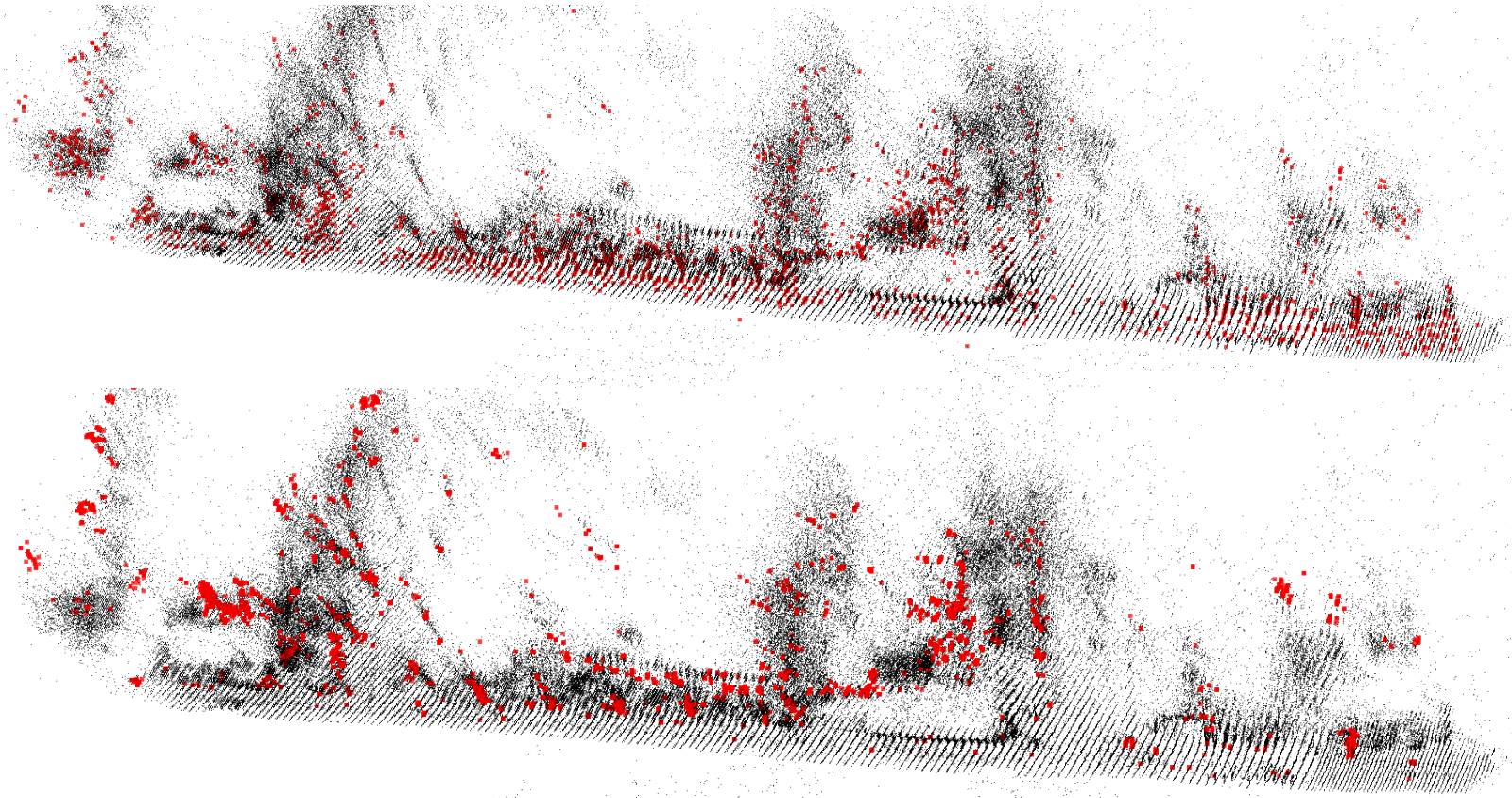}  \\
		{\small(a) slice 3} & {\small(b) slice 11}
	\end{tabular}	
	\vspace{-3mm}
	\caption{Large-scale point selection results. The upper row is the results from ILP (map) and the lower row is ours with a 0.1 score threshold. The black points are the map 3D points before sparsification and the red points are the selected points. Our method selects points on static structures, such as building walls, utility poles, and tree stems and avoids foliage that changes across seasons.}
	\label{fig:result_scores}
 \end{figure*}

\begin{figure*} [h]
	\centering
	\begin{tabular}{c c c c }
		\includegraphics[width=0.19\textwidth]{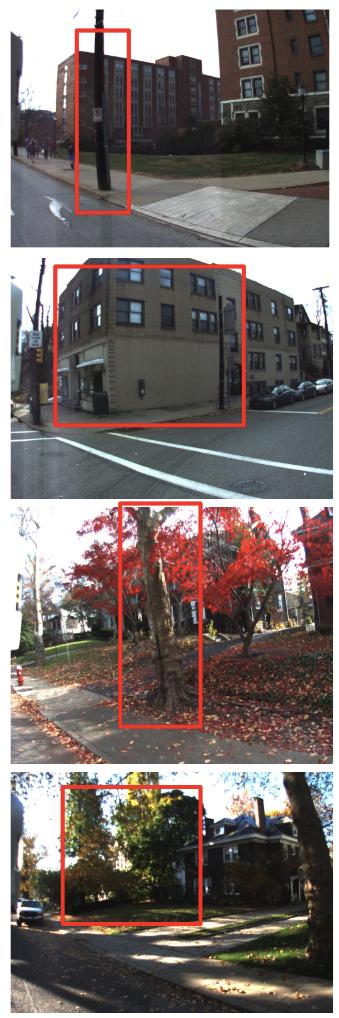}  & \includegraphics[width=0.19\textwidth]{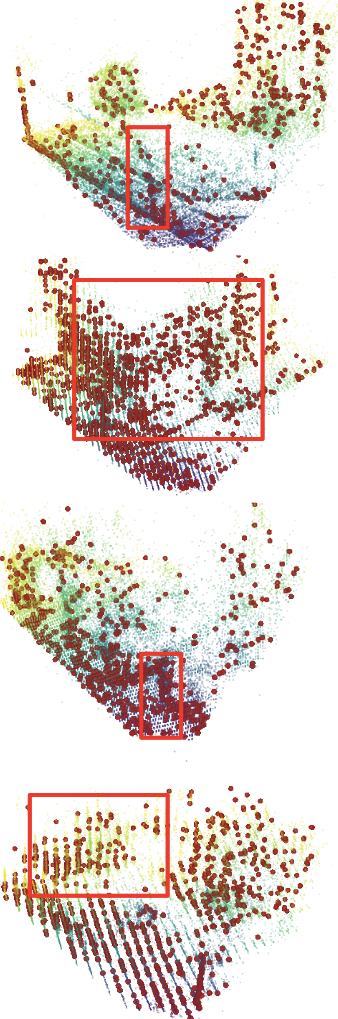} &\includegraphics[width=0.19\textwidth]{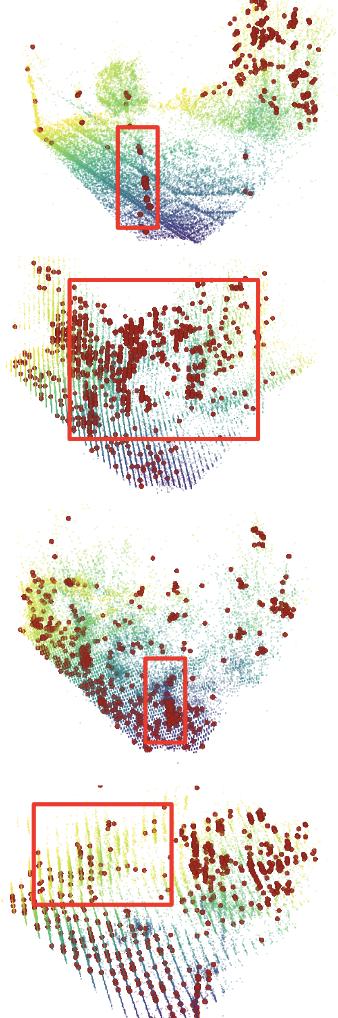}&\includegraphics[width=0.19\textwidth]{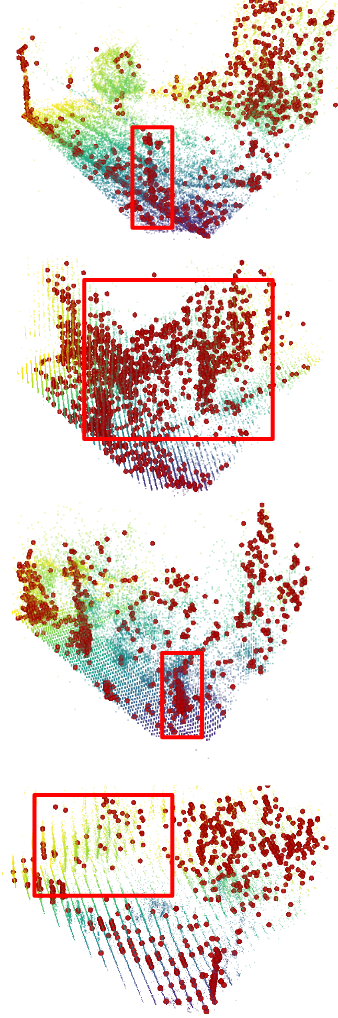}  \\
		{\small(a) images} & {\small(b) ILP (map)} & {\small(c) GATConv (ours)} & {\small(d) ILP (query)}\\
	\end{tabular}	
	\vspace{-2mm}
	\caption{ Qualitative visualizations. The camera positions are at the bottoms of the point cloud visualizations (b)(c)(d). The corresponding parts in each row are labels by red boxes. Overall, we observe that the point selection of ILP (map) is less discriminative in selecting static points than ILP (query) and ours. We compared the cases with similar numbers of key points so the total 3D point number varies. 
	\vspace{-3mm}
	}
	\label{fig:result_imgs}

\end{figure*}

Overall, our proposed approach outperformed the ILP (map) baseline in all the testing slices by achieving higher localization recall (success rate) under the same map sizes, as shown in  Tab.~\ref{tb:results_recalls} and Fig.~\ref{fig:result_curves}. Qualitatively, we observed that compared with the ILP (map) baseline, the proposed method selects map points on static structures that are more useful for query set localization, as in Fig.~\ref{fig:result_scores} and Fig.~\ref{fig:result_imgs}. \\

\noindent\textbf{Network structures}. We also compared the following configurations for the $g_2$ GNN layer: GraphConv~\cite{kipf2017}, SAGEConv (with mean aggregation function)~\cite{hamilton2018}, and GATConv (with $H=4$)~\cite{velickovic2018}. The compared networks had the same feature dimensions and the LeakyReLU ($slope=0.1$) activations. 
Our results showed GATConv outperformed GraphConv and SAGEConv significantly in terms of not only localization recall (Tab.~\ref{tb:results_recalls}) under the same map sizes, but also  classification performance with respect to ILP (query) as shown in Fig.~\ref{fig:result_curves}(b). \\

\begin{figure} 
	\centering
	\includegraphics[width=0.48\textwidth]{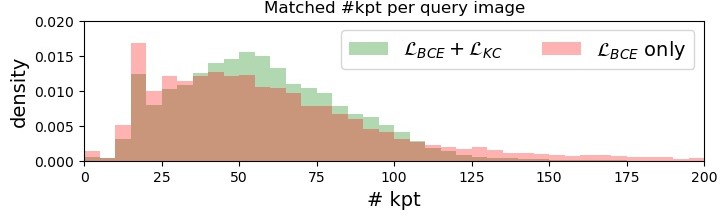} 
	\vspace{-2mm}
	\caption{The density histogram of 2D-3D matching number for each testing query image during localization. After applying $\mathcal{L}_{KC}$ we observed less images with extreme number of matches, which is preferred for consistent localization performance under a map size budget. Both histograms are generated under the same map size budget (total \#kpt is $\sim6.3\times10^5$).}
	\label{fig:kpt_dist}
	\vspace{-4mm}
\end{figure}

\begin{table*}[ht]
\centering
\small
\caption{Average recall under different map sizes. For each slice (a sequence in The Extended CMU Seasons dataset), we linearly interpolated the recall curves to obtain the recall numbers under the same number of key point descriptors, and computed the average recalls with respect to the number of images. Three recall thresholds were compared. The recall number represents the ratio of image samples with localization pose errors less than the corresponding recall threshold. As a reference, the average key point number before sparsification is $~\sim2.8\times10^6$. The detailed map graph statistics and the full recall curves are in the supplementary material.
}
\begin{tabular}{c|c c c c|c c c c|c c c c}
\toprule				
Recall threshold& \multicolumn{4}{c|}{\bf{0.25m, 2.0$^\circ$}}	 	&	\multicolumn{4}{c|}{\bf{0.5m, 5.0$^\circ$}}	&	\multicolumn{4}{c}{\bf{5.0m, 10.0$^\circ$}}	\\
\midrule
Avg. map size ($10^4$ \#kpts)&	3	&	5	&	10	&	20	&	3	&	5	&	10	&	20	&	3	&	5	&	10	&	20\\
\midrule
Random	&	0.07	&	0.18	&	0.41	&	0.59	&	0.07	&	0.20	&	0.44	&	0.63	&	0.09	&	0.23	&	0.49	&	0.70	\\
ILP (map)	&	0.15	&	0.31	&	0.53	&	0.64	&	0.19	&	0.36	&	0.59	&	0.69	&	0.25	&	0.43	&	0.66	&	0.76	\\
\midrule
GraphConv	&	0.31	&	0.48	&	0.64	&	\bf{0.73}	&	0.34	&	0.52	&	0.69	&	0.77	&	0.39	&	0.58	&	0.76	&	0.85	\\
SAGEConv	&	0.27	&	0.42	&	0.58	&	0.68	&	0.30	&	0.46	&	0.62	&	0.72	&	0.34	&	0.51	&	0.68	&	0.79	\\
GATConv (ours)	&	\bf{0.35}	&	\bf{0.52}	&	\bf{0.67}	&	\bf{0.73}	&	\bf{0.40}	&	\bf{0.57}	&	\bf{0.72}	&	\bf{0.78}	&	\bf{0.46}	&	\bf{0.64}	&	\bf{0.80}	&	\bf{0.86}	\\
\midrule
GATConv ($\mathcal{L}_{BCE}$ only)	&	0.25	&	0.38	&	0.53	&	0.65	&	0.28	&	0.42	&	0.57	&	0.70	&	0.32	&	0.47	&	0.64	&	0.77	\\
GATConv ($\mathcal{L}_{KC}$ only)	&0.09	&0.23	&0.42	&0.60	&0.10	&0.25	&0.45	&0.64	&0.12	&0.29	&0.52	&0.71	\\
\midrule
\color{gray}ILP (query)	&	\color{gray}0.24	&	\color{gray}0.46	&\color{gray}	\color{gray}0.69	&	\color{gray}0.80	&	\color{gray}0.30	&	\color{gray}0.53	&	\color{gray}0.75	&	\color{gray}0.85	&	\color{gray}0.38	&	\color{gray}0.60	&	\color{gray}0.83	&	\color{gray}0.92	\\
\bottomrule    									
\end{tabular}
\label{tb:results_recalls}
\end{table*}

\vspace{-2mm}
\noindent\textbf{Training losses}. Finally, the network trained without either $\mathcal{L}_{BCE}$ or $\mathcal{L}_{KC}$ performed worse than the one with combined loss, as shown in Tab.~\ref{tb:results_recalls} and Fig.~\ref{fig:result_curves}(b). The $\mathcal{L}_{BCE}$ was only trained in the training area, since no labels are available in the testing area. The $\mathcal{L}_{KC}$ were trained with the whole input map graph (which covers both the training and testing areas). Interestingly, although the $\mathcal{L}_{BCE}$-only configuration got the lowest training $\mathcal{L}_{BCE}$, adding $\mathcal{L}_{KC}$ improved the classification performance in the test set. 
We further observed that when localizing testing queries, the map sparsified with $\mathcal{L}_{KC}$ obtained less extreme numbers of matched key points (Fig.~\ref{fig:kpt_dist}).
This is favorable because each query obtained enough matches, but not too many that caused a waste in map storage. \\


\vspace{-6mm}
\section{Discussion and Limitations}
The heterogeneous graph used in this work is so flexible that it is easy to include more information as additional node or edge features. This implies a great potential for future works. Choices of additional information include timestamps (for capturing periodic environmental change) or the data from other sensors. It is also easy to apply other training losses to sparsify the map for different tasks other than conventional localization. Furthermore, we observed that certain objects, such as buildings and utility poles. are more likely to get higher scores. This implies the possibility of using semantic labels to assist point score prediction. It is also worth mentioning that the heterogeneous GNN framework can potentially be applied to other practical graphs, such as the factor graph for in SLAM. Comparing the GNN-based method with the existing factor graph sparsification works \cite{cadena2016} is another interesting future direction. On the other hand, one important factor affecting the result is the point sampling strategy. Given the same set of predicted scores, different point selection strategies would lead to different performance. In our system, we used simple random down-sampling and a score threshold that achieved outstanding performance, but exploring different point sampling strategies can be an interesting future work.

As for limitations, typically the key in map sparsification is to compress a map of a given scene, thus the generalization to an unseen scene has not been our focus. For the K-Cover setup to work, the camera trajectories at query time should be a subset of the camera trajectories in the map. This applies to ours and the related works. Besides, we only focused on removing points from an existing map, so the result is limited by the localization performance on the raw map. How to add/merge new information to the map is also worth exploring in the future. Finally, naive data splits (by camera and by slice) is used in our experiments, but in practice it is better to minimize the training set size to reduce the map update workload.
\vspace{-2mm}
\section{Conclusion}
\vspace{-2mm}
In conclusion, we proposed a heterogeneous GNN for visual map sparsification and proved its effectiveness in real-world environment. This work opens a new avenue for applying the abundant GNN related techniques to SfM applications. Our future work would be map sparsification for multi-sensor maps and more map graph representations.

\section*{Acknowledgement}
Ming-Fang Chang is supported by the CMU Argo AI Center for Autonomous Vehicle Research. We also thank Tianwei Shen and dear labmates in CMU and Meta for providing valuable discussion and suggestions.

\appendix

\section{Details of Conventional ILP Method}

Here we describe the details of the ILP method we used to generate $L_{gt}$ and as baselines~\cite{park2013,dymczyk2015, lynen2019d}. Given $N_p$ 3D points and $N_m$ key frame images in the map, letting $\mathbf{x}$ be the binary point selection vector, we can formulate the following ILP problem:
\begin{equation}
	\begin{aligned}	
				minimize~& \mathbf{q}^T \mathbf{x} + \lambda \mathbf{1}^T \mathbf{\zeta} \\
				s.t.&~\mathbf{A}\mathbf{x} + \mathbf{\zeta} \geq b\mathbf{1}\\
				& \sum_{i=1}^{N_p} \mathbf{x}_i = n_{desired}\\
				&\mathbf{x} \in \{0,1\}^{N_p} \\
				& \zeta \in \{\{0\} \cup \mathbb{Z}^{+}\}^{N_m} ,
			\end{aligned}
	\label{eq:ilp}
\end{equation}
where $\mathbf{q}$ is an assigned weighting vector, $\mathbf{A} \in \mathbb{R}^{N_m \times N_p}$ is the visibility matrix, $\mathbf{\zeta}$ is the slack variable, $b$ is a tunable variable indicating the desired minimum number of observable 3D points for each map key frame ($b=30$ was used in the paper as in~\cite{dymczyk2015}), and $n_{desired}$ is the desired total 3D point number. The weighting vector $\mathbf{q}$ is computed from the observation count (the number of times a 3D point is matched by a 2D key point in the map building history) of each 3D point. Let $c_i$ denote the observation count of a 3D point with index $i$, the corresponding $q_i$ for this point is assigned as $max({c_1, c_2,\dots, c_{N_p}})-c_i$ as in~\cite{park2013}. The idea is to assume the well-observed points in the past to play a more important role in future localization, and decrease the weights of these points in the minimization cost. In a dynamic environment where the physical structures of the world changes, a point that was observed for more times in the past might not exist or as important in the future, even if it has a robust feature descriptor.

\section{Full Recall Curves}

\begin{figure*} 
	\centering
\begin{subfigure}{0.7\textwidth}
	\centering
	\includegraphics[width=0.99\textwidth]{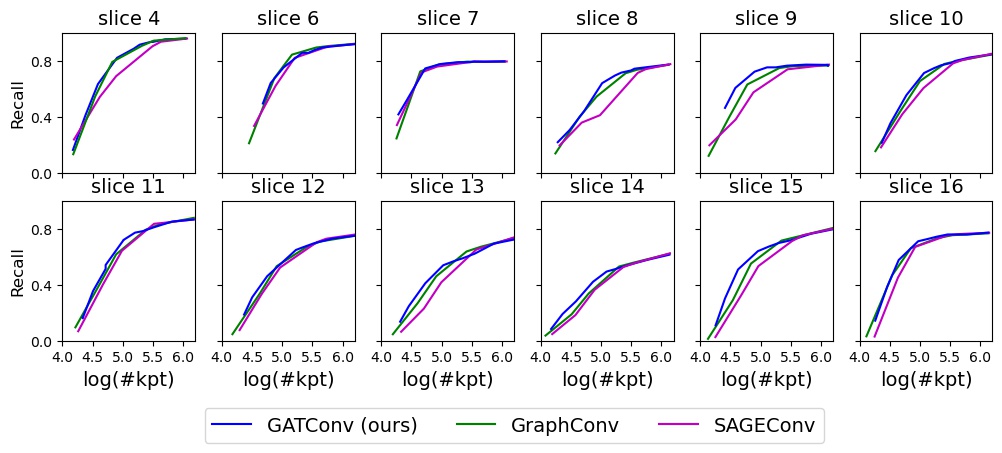} 
	\caption{The full recall curves of different $g_2$ layers.}
	\label{fig:conv_type}
\end{subfigure}
\begin{subfigure}{0.7\textwidth}
	\centering
	\includegraphics[width=0.99\textwidth]{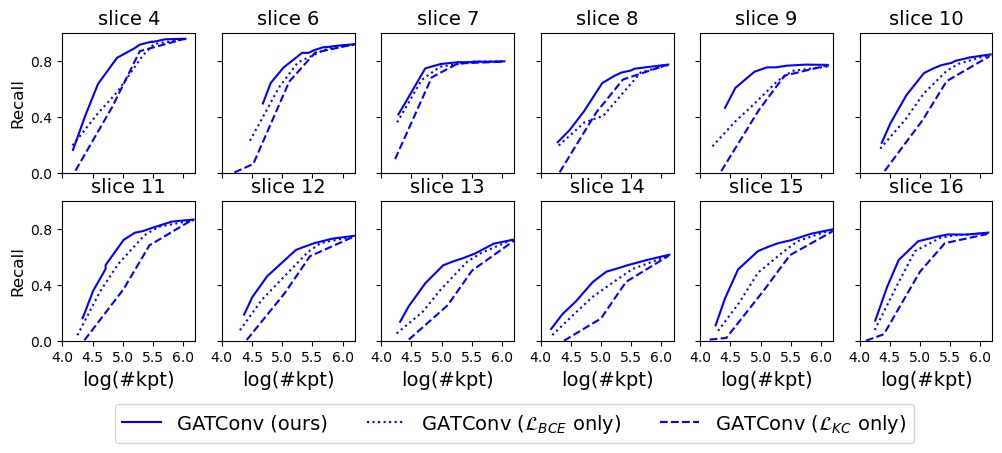} 
	\caption{The full recall curves of using different losses.}
	\label{fig:losses}
\end{subfigure}
\begin{subfigure}{0.7\textwidth}
	\centering
	\includegraphics[width=0.99\textwidth]{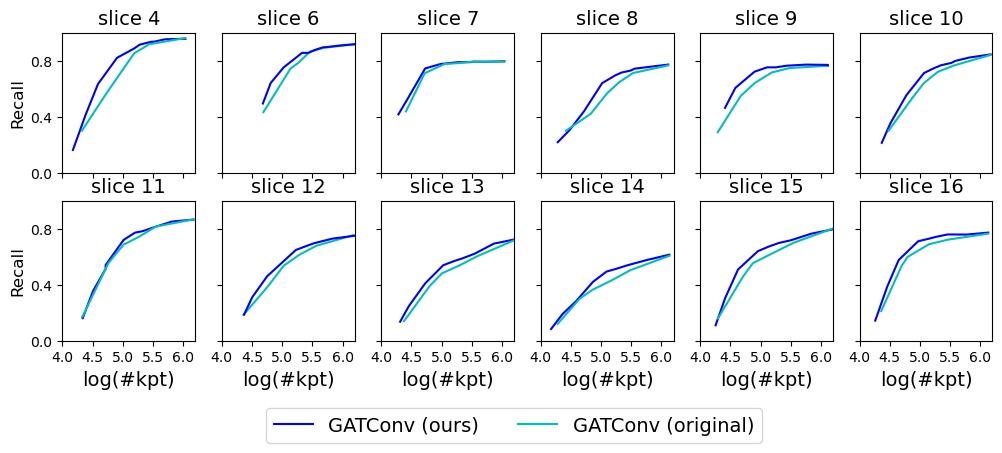} 
	\caption{The full recall curves of different GATConv versions.}
	\label{fig:gatconv_versions}
\end{subfigure}
\end{figure*}

We present the full recall curves for all the methods compared, for which we only provided linearly interpolated and averaged numbers in Tab.~\ref{tb:results_recalls} due to space limitation. In Fig.~\ref{fig:conv_type} we show the full recall curves of $g_2$ layer using GATConv, GraphConv, and SAGEConv. In Fig.~\ref{fig:losses} we show the full curves of the proposed combined loss and without either $\mathcal{L}_{KC}$ or $\mathcal{L}_{BCE}$. One small difference between our GATConv and the original version in~\cite{velickovic2018} is the way of merging multi-head attentions. In~\cite{velickovic2018} the authors computed the mean of all the heads and added an non-linear layer, while we performed simple summation. Empirically we found simple summation outperforms the original version, as shown in Fig.~\ref{fig:gatconv_versions}.

\section{Map Graph Statistics}

To better describe the scale of the localization problem we are solving, we listed the number of nodes in each testing map graph and the corresponding processing time (on CPU) for the proposed GNN to process the whole map graph and generate scores. Note that number of images in the map graph is slightly less than Sec.~\ref{sec:eval} because some images were discarded during the mapping process. To perform efficient training on these large map graphs, we iterated through extracted local map graphs during training on GPU, and computed the scores of the whole map graphs at once on CPU with trained weights.

\begin{figure*} 
	\centering
	\includegraphics[width=0.8\textwidth]{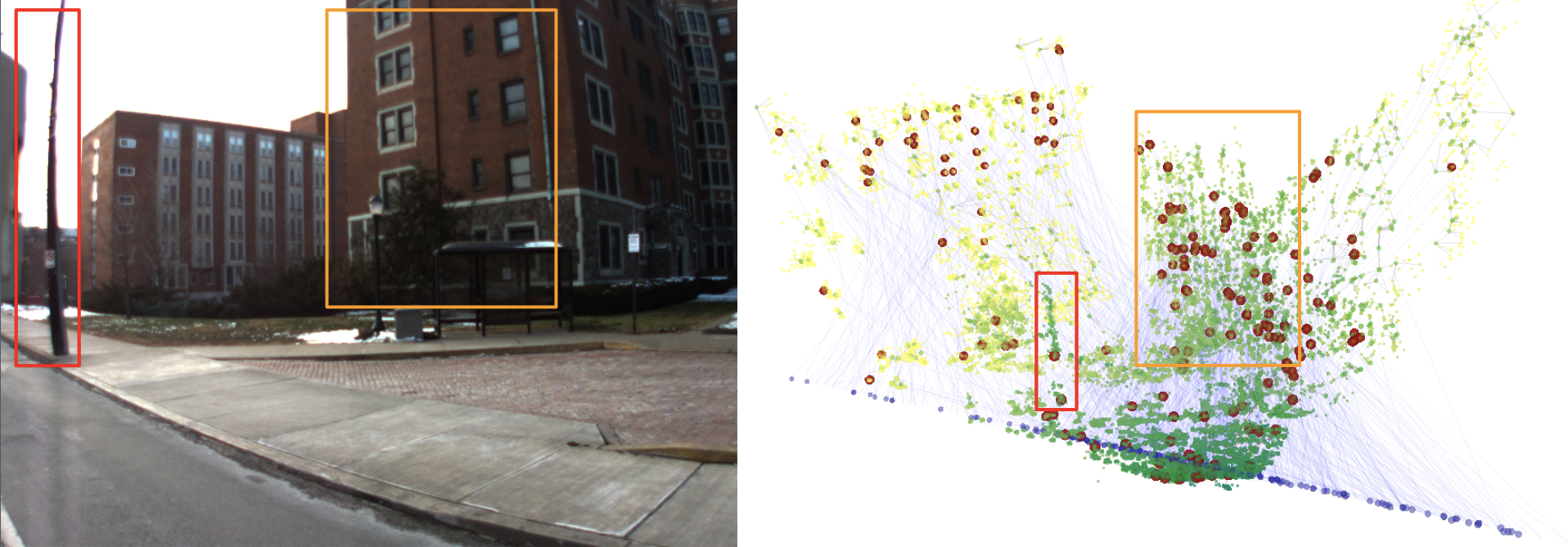} 
	\includegraphics[width=0.8\textwidth]{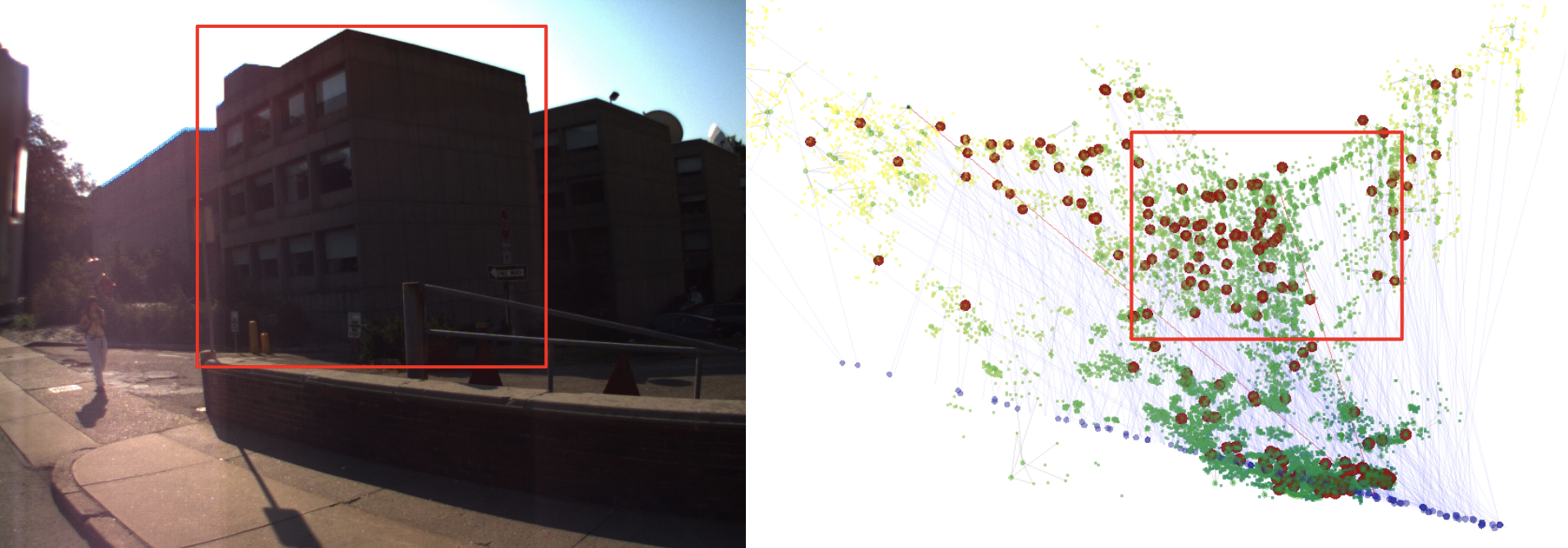} 
	\caption{More visualizations of our point selection, where the $\mathcal{V}_p$, $\mathcal{V}_m$, $\mathcal{E}_v$, $\mathcal{E}_n$ are shown by green/yellow dots, blue dots, blue lines, and gray lines. The edges were downsampled for better visualization clarity. Assigning $n_{desired}=200$, our method selects the red dots from the green dots (the $\mathcal{V}_p$ connected to the key points in the sampled map image). On the left, we show the corresponding image to the local map graph. The corresponding parts are shown by red and orange boxes. }
	\label{fig:visual}
\end{figure*}

\begin{table*}[ht]
	\centering
	\caption{Test map graph sizes stats and the processing time. The graph size is represented by the number of nodes and edges. Each $\mathcal{V}_k$ carries an R2D2 descriptor with dimension 128.}
	\begin{tabular}{c|c c c c c c c c}
		
		\toprule		
		slice	&	time (ms)	&	\# $\mathcal{V}_p$	&		\# $\mathcal{V}_m$	&	\# $\mathcal{V}_k$	&	\# $\mathcal{E}_c$	&	\# $\mathcal{E}_k$	&	\# $\mathcal{E}_n$\\	
		\midrule
		4	&	3,188.20	&	326,797 	&	1,260 	&	2,072,834 	&	2,072,834 	&	2,072,834 	&	3,267,970 	\\
		6	&	5,261.06	&	508,844 	&	1,642 	&	3,632,027 	&	3,632,027 	&	3,632,027 	&	5,088,440 	\\
		7	&	2,982.98	&	320,258 	&	1,196 	&	2,034,125 	&	2,034,125 	&	2,034,125 	&	3,202,580 	\\
		8	&	3,440.97	&	377,380 	&	1,423 	&	2,465,051 	&	2,465,051 	&	2,465,051 	&	3,773,800 	\\
		9	&	3,563.63	&	387,988 	&	1,154 	&	2,531,802 	&	2,531,802 	&	2,531,802 	&	3,879,880 	\\
		10	&	4,023.90	&	433,398 	&	1,356 	&	2,830,219 	&	2,830,219 	&	2,830,219 	&	4,333,980 	\\
		11	&	3,930.64	&	421,625 	&	1,394 	&	2,859,143 	&	2,859,143 	&	2,859,143 	&	4,216,250 	\\
		12	&	4,143.14	&	439,425 	&	1,454 	&	2,960,501 	&	2,960,501 	&	2,960,501 	&	4,394,250 	\\
		13	&	4,395.27	&	456,772 	&	1,409 	&	3,275,294 	&	3,275,294 	&	3,275,294 	&	4,567,720 	\\
		14	&	3,806.34	&	427,022 	&	1,392 	&	2,629,698 	&	2,629,698 	&	2,629,698 	&	4,270,220 	\\
		15	&	4,474.99	&	469,165 	&	1,418 	&	3,156,230 	&	3,156,230 	&	3,156,230 	&	4,691,650 	\\
		16	&	3,527.82	&	370,006 	&	1,319 	&	2,617,829 	&	2,617,829 	&	2,617,829 	&	3,700,060 	\\
		\bottomrule    									
	\end{tabular}
	\label{tb:proc_time}
\end{table*}

\section{More Map Graph Visualization}

\begin{figure*} 
	\centering
	\includegraphics[width=0.8\textwidth]{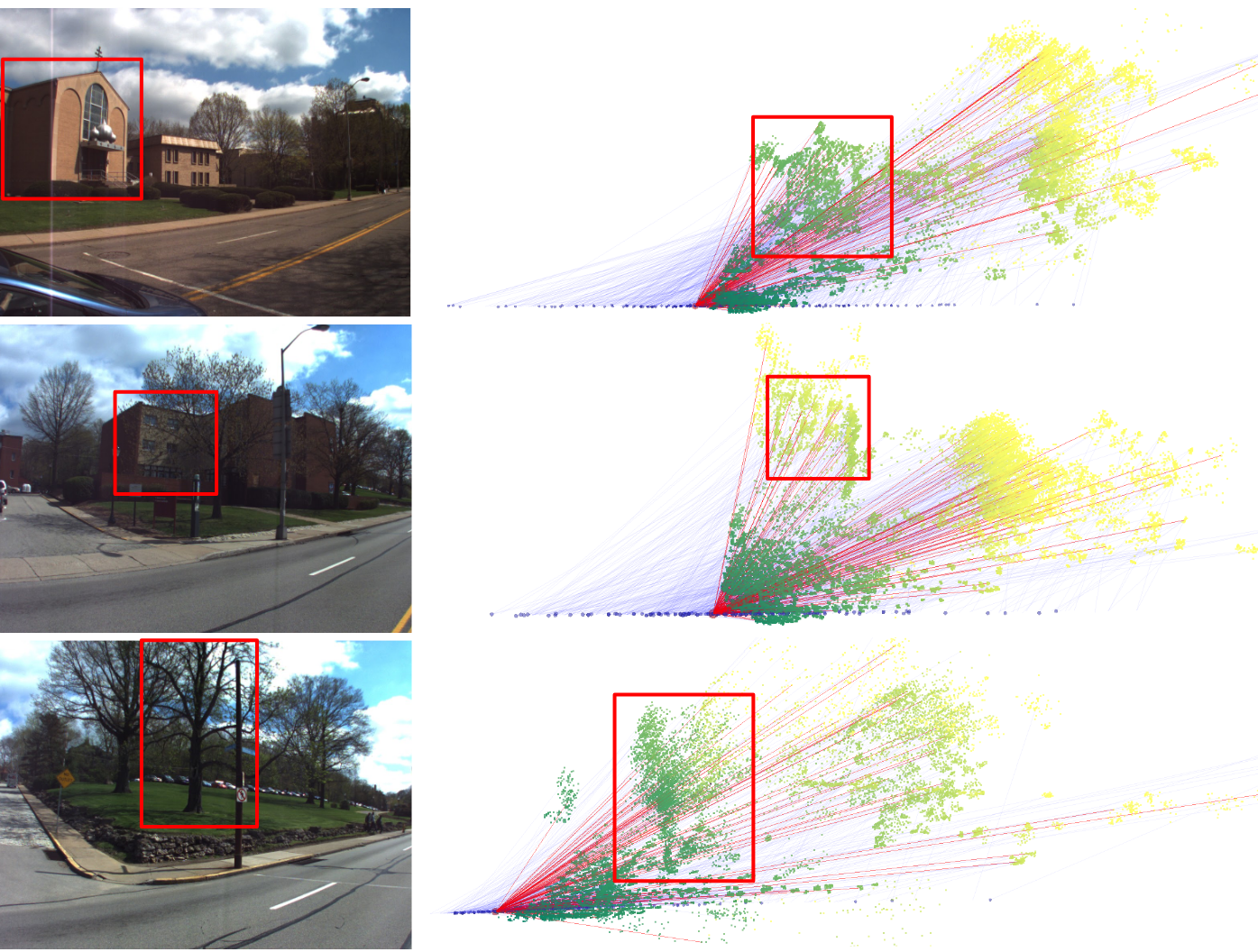} 
	\includegraphics[width=0.8\textwidth]{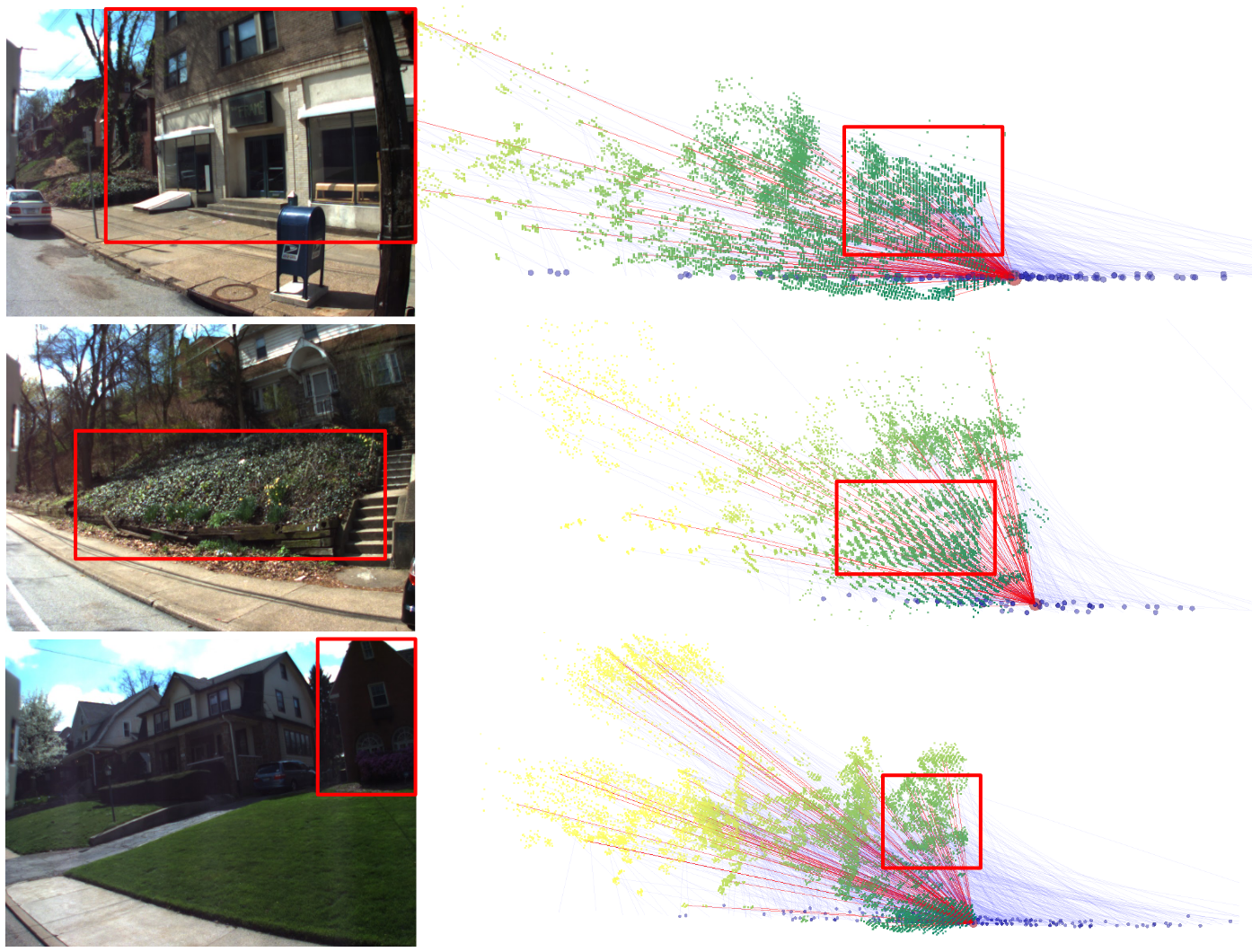} 
	\caption{More local graph visualizations. The upper three graphs were extracted from camera 0, and the lower three graphs were extracted from camera 1. The corresponding parts of the images and the graphs are shown by red boxes. The $\mathcal{V}_p$, $\mathcal{V}_m$, $\mathcal{E}_v$ from the map are shown by green/yellow dots, blue dots, and blue lines. The $\mathcal{E}_v$ connected to the sampled $\mathcal{V}_m$ are shown by red lines. These map graphs provide all the information needed to predict scores for the $\mathcal{V}_p$ connected by the red $\mathcal{E}_v$ set.}
	\label{fig:submaps}
\end{figure*}

Here we visualize more samples of local map graphs in Fig.~\ref{fig:submaps}. Each of the local map graph was extracted by first sampling a map image, and trace the connected edges to acquire all the information needed. Besides the map graphs, we can also observe the variety of objects in this dataset from Fig.~\ref{fig:submaps}.


{\small
\bibliographystyle{ieee_fullname}
\bibliography{egbib}

\begin{thebibliography}{10}\itemsep=-1pt

\bibitem{iccv2021challenge}
Long-term visual localization challenges in iccv 2021.
\newblock \url{https://sites.google.com/view/ltvl2021/challenges}.
\newblock Accessed: 2022-03-09.

\bibitem{Badino2011}
Hernan Badino, Daniel Huber, and Takeo Kanade.
\newblock {The CMU Visual Localization Data Set}.
\newblock \url{http://3dvis.ri.cmu.edu/data-sets/localization}, 2011.

\bibitem{cadena2016}
Cesar Cadena, Luca Carlone, Henry Carrillo, Yasir Latif, Davide Scaramuzza,
  Jose Neira, Ian Reidand, and John~J. Leonard.
\newblock Past, present, and future of simultaneous localization and mapping:
  Toward the robust-perception age.
\newblock 2016.

\bibitem{camposeco2019}
Federico Camposeco, Andrea Cohen, Marc Pollefeys, and Torsten Sattler.
\newblock Hybrid scene compression for visual localization.
\newblock In {\em IEEE Conf. Comput. Vis. Pattern Recog.}, 2019.

\bibitem{cao2014}
Song Cao and Noah Snavely.
\newblock Minimal {{Scene Descriptions}} from {{Structure}} from {{Motion
  Models}}.
\newblock In {\em IEEE Conf. Comput. Vis. Pattern Recog.}, 2014.

\bibitem{doan2021learning}
Anh-Dzung Doan, Daniyar Turmukhambetov, Yasir Latif, Tat-Jun Chin, and Soohyun
  Bae.
\newblock {Learning to Predict Repeatability of Interest Points}.
\newblock In {\em IEEE Int. Conf. Robotics and Automation}, 2021.

\bibitem{dymczyk2015}
Marcin Dymczyk, Simon Lynen, Michael Bosse, and Roland Siegwart.
\newblock {Keep It Brief: Scalable Creation of Compressed Localization Maps}.
\newblock In {\em IEEE/RSJ Int. Conf. Intell. Robots and Syst.}, 2015.

\bibitem{dymczyk2016will}
Marcin Dymczyk, Elena Stumm, Juan Nieto, Roland Siegwart, and Igor
  Gilitschenski.
\newblock {Will it last? Learning stable features for long-term visual
  localization}.
\newblock In {\em IEEE Int. Conf. on 3D Vis.}, 2016.

\bibitem{gurobioptimizationllc2021}
{Gurobi Optimization, LLC}.
\newblock Gurobi - {{The Fastest Solver}}.
\newblock https://www.gurobi.com/, 2021.

\bibitem{hamilton2018}
William~L. Hamilton, Rex Ying, and Jure Leskovec.
\newblock Inductive {{Representation Learning}} on {{Large Graphs}}.
\newblock In {\em Conf. Neural Inform. Process. Syst.}, 2017.

\bibitem{humenberger2020}
Martin Humenberger, Yohann Cabon, Nicolas Guerin, Julien Morat, J{\'e}r{\^o}me
  Revaud, Philippe Rerole, No{\'e} Pion, Cesar {de Souza}, Vincent Leroy, and
  Gabriela Csurka.
\newblock Robust {{Image Retrieval}}-based {{Visual Localization}} using
  {{Kapture}}.
\newblock {\em arXiv:2007.13867 [cs]}, 2020.

\bibitem{kipf2017}
Thomas~N. Kipf and Max Welling.
\newblock Semi-{{Supervised Classification}} with {{Graph Convolutional
  Networks}}.
\newblock In {\em Int. Conf. Learn. Represent.}, 2017.

\bibitem{Luthardt2018}
Stefan Luthardt, Volker Willert, and Jürgen Adamy.
\newblock {LLama-SLAM}: Learning high-quality visual landmarks for long-term
  mapping and localization.
\newblock In {\em IEEE Int. Conf. Intell. Transportation Syst.}, 2018.

\bibitem{lynen2019d}
Simon Lynen, Bernhard Zeisl, Dror Aiger, Michael Bosse, Joel Hesch, Marc
  Pollefeys, Roland Siegwart, and Torsten Sattler.
\newblock Large-scale, real-time visual-inertial localization revisited.
\newblock {\em Int. J. of Robotics Res.}, 2020.

\bibitem{mera-trujillo2020a}
Marcela {Mera-Trujillo}, Benjamin Smith, and Victor Fragoso.
\newblock Efficient {{Scene Compression}} for {{Visual}}-based
  {{Localization}}.
\newblock In {\em IEEE Int. Conf. on 3D Vis.}, 2020.

\bibitem{muhlfellner2016}
Peter Mühlfellner, Mathias Bürki, M. Bosse, W. Derendarz, Roland Philippsen,
  and P. Furgale.
\newblock Summary maps for lifelong visual localization.
\newblock In {\em J. of Field Robotics}, 2016.

\bibitem{park2013}
Hyun~Soo Park, Yu Wang, Eriko Nurvitadhi, James~C. Hoe, Yaser Sheikh, and Mei
  Chen.
\newblock {{3D Point Cloud Reduction Using Mixed}}-{{Integer Quadratic
  Programming}}.
\newblock In {\em IEEE Conf. Comput. Vis. Pattern Recog. Worksh.}, 2013.

\bibitem{piasco2019}
Nathan Piasco, Désiré Sidibé, Valérie Gouet-Brunet, and Cedric Demonceaux.
\newblock {Learning Scene Geometry for Visual Localization in Challenging
  Conditions}.
\newblock In {\em IEEE Int. Conf. Robotics and Automation}, 2019.

\bibitem{revaud2019}
Jerome Revaud, Philippe Weinzaepfel, C{\'e}sar De~Souza, Noe Pion, Gabriela
  Csurka, Yohann Cabon, and Martin Humenberger.
\newblock {{R2D2}}: Repeatable and {{Reliable Detector}} and {{Descriptor}}.
\newblock In {\em Conf. Neural Inform. Process. Syst.}, 2019.

\bibitem{sarlin2020}
Paul-Edouard Sarlin, Daniel DeTone, Tomasz Malisiewicz, and Andrew Rabinovich.
\newblock {{SuperGlue}}: Learning {{Feature Matching With Graph Neural
  Networks}}.
\newblock In {\em IEEE Conf. Comput. Vis. Pattern Recog.}, 2020.

\bibitem{sarlin2021back}
Paul-Edouard Sarlin, Ajaykumar Unagar, Mans Larsson, Hugo Germain, Carl Toft,
  Viktor Larsson, Marc Pollefeys, Vincent Lepetit, Lars Hammarstrand, Fredrik
  Kahl, et~al.
\newblock {Back to the Feature: Learning Robust Camera Localization from Pixels
  to Pose}.
\newblock In {\em IEEE Conf. Comput. Vis. Pattern Recog.}, 2021.

\bibitem{sattler2018a}
Torsten Sattler, Will Maddern, Carl Toft, Akihiko Torii, Lars Hammarstrand,
  Erik Stenborg, Daniel Safari, Masatoshi Okutomi, Marc Pollefeys, Josef Sivic,
  Fredrik Kahl, and Tomas Pajdla.
\newblock Benchmarking {{6DOF Outdoor Visual Localization}} in {{Changing
  Conditions}}.
\newblock In {\em IEEE Conf. Comput. Vis. Pattern Recog.}, 2018.

\bibitem{shen2021}
Yan Shen, Zhang Maojun, {Lai, Shiming}, Liu Yu, and Peng Yang.
\newblock {Image Retrieval for {{Structure}}-from-{{Motion}} via {{Graph
  Convolutional Network}}}.
\newblock {\em Inform. Sci.}, 2021.

\bibitem{tian2019sosnet}
Yurun Tian, Xin Yu, Bin Fan, Fuchao Wu, Huub Heijnen, and Vassileios Balntas.
\newblock {SoSnet: Second Order Similarity Regularization for Local Descriptor
  Learning}.
\newblock In {\em IEEE Conf. Comput. Vis. Pattern Recog.}, 2019.

\bibitem{toft2020long}
Carl Toft, Will Maddern, Akihiko Torii, Lars Hammarstrand, Erik Stenborg,
  Daniel Safari, Masatoshi Okutomi, Marc Pollefeys, Josef Sivic, Tomas Pajdla,
  et~al.
\newblock {Long-Term Visual Localization Revisited}.
\newblock {\em IEEE Trans. Pattern Anal. Mach. Intell.}, 2020.

\bibitem{toft12017}
Carl Toft, Carl Olsson, and Fredrik Kahl.
\newblock {Long-term 3D Localization and Pose from Semantic Labellings}.
\newblock In {\em Int. Conf. Comput. Vis. Worksh.}, 2017.

\bibitem{toft12018}
Carl Toft, Erik Stenborg, Lars Hammarstrand, Lucas Brynte, Marc Pollefeys,
  Torsten Sattler, and Fredrik Kahl.
\newblock {Semantic Match Consistency for Long-Term Visual Localization}.
\newblock In {\em Eur. Conf. Comput. Vis.}, 2018.

\bibitem{vaswani2017}
Ashish Vaswani, Noam Shazeer, Niki Parmar, Jakob Uszkoreit, Llion Jones,
  Aidan~N. Gomez, Lukasz Kaiser, and Illia Polosukhin.
\newblock Attention {{Is All You Need}}.
\newblock In {\em Conf. Neural Inform. Process. Syst.}, 2017.

\bibitem{velickovic2018}
Petar Veli{\v c}kovi{\'c}, Guillem Cucurull, Arantxa Casanova, Adriana Romero,
  Pietro Li{\`o}, and Yoshua Bengio.
\newblock Graph {{Attention Networks}}.
\newblock In {\em Int. Conf. Learn. Represent.}, 2018.

\bibitem{wang2020c}
Minjie Wang, Da Zheng, Zihao Ye, Quan Gan, Mufei Li, Xiang Song, Jinjing Zhou,
  Chao Ma, Lingfan Yu, Yu Gai, Tianjun Xiao, Tong He, George Karypis, Jinyang
  Li, and Zheng Zhang.
\newblock Deep {{Graph Library}}: A {{Graph}}-{{Centric}},
  {{Highly}}-{{Performant Package}} for {{Graph Neural Networks}}.
\newblock {\em arXiv:1909.01315 [cs, stat]}, 2020.

\bibitem{wang2021a}
Xiao Wang, Houye Ji, Chuan Shi, Bai Wang, Peng Cui, P. Yu, and Yanfang Ye.
\newblock Heterogeneous {{Graph Attention Network}}.
\newblock In {\em The World Wide Web Conf.}, 2021.

\end{thebibliography}
}

\end{document}